\documentclass{article} 
\usepackage{iclr2027_conference,times}

\iclrfinaltrue

\usepackage{amsmath,amsfonts,bm}

\def\eqref#1{equation~\ref{#1}}

\def\1{\bm{1}}

\DeclareMathAlphabet{\mathsfit}{\encodingdefault}{\sfdefault}{m}{sl}
\SetMathAlphabet{\mathsfit}{bold}{\encodingdefault}{\sfdefault}{bx}{n}

\usepackage{hyperref}
\usepackage{url}
\usepackage{graphicx}
\usepackage{booktabs}
\usepackage{multirow}
\usepackage{capt-of}
\usepackage{wrapfig}
\usepackage{placeins}
\usepackage{amsmath}
\usepackage{amssymb}
\usepackage{xcolor}

\newcommand{\Sim}{\textsc{Sim}\%}
\newcommand{\Miss}{\textsc{Miss}\%}
\newcommand{\method}{\textsc{TACR}}
\newcommand{\twuq}{\textsc{TW-UQ}}
\newcommand{\gh}{\hat{g}}
\newcommand{\yh}{\hat{y}}
\newcommand{\qh}{\hat{q}}
\newcommand{\prox}{\mathrm{prox}}

\newcommand{\blue}[1]{#1}

\title{Threshold-Aware Conformal Routing}

\author{Shiwei Tan\thanks{Work done during an internship at Siemens.} \\
Rutgers University \\
\texttt{shiwei.tan@rutgers.edu} \\
\And
Huzefa Rangwala \\
Siemens \\
\texttt{huzefa.rangwala@siemens.com}
\And
Danielle C. Maddix \\
Siemens \\
\texttt{danielle.maddix-robinson@siemens.com} \\
}

\begin{document}

\maketitle
\lhead{Preprint}

\begin{abstract}

High-fidelity simulations are essential to scientific and engineering design, but can be expensive to run repeatedly. Learned surrogates offer a faster alternative, yet their higher errors may alter downstream decisions. This accuracy-speed tradeoff creates a need to determine whether a surrogate can be used or the full simulator remains necessary. We study decisions determined by whether a scalar quantity of interest lies above or below a fixed threshold. For each input, we use the surrogate when its conformal interval lies entirely on one side of the threshold and route the input to simulation when the interval intersects it. Standard conformal prediction constructs intervals without reference to the downstream decision threshold: even a narrow interval near the threshold can cross it and trigger simulation, whereas a wider interval farther away can remain entirely on one side and require no simulation.
We introduce Threshold-Aware Conformal Routing (\method), which learns an input-dependent scale using a threshold-aware objective that concentrates interval tightness near the decision boundary. Exact split-conformal calibration on held-out data preserves distribution-free marginal coverage, which also upper-bounds the probability of an incorrect threshold decision that is not routed. Across various scientific and engineering datasets, \method{} reduces simulator deferrals by 14--75\% relative to standard conformal prediction at the same coverage target. Against a variant without threshold-local weighting but with similar predictor accuracy, \method{} further reduces deferrals by 10--24\% on four datasets. These results show that optimizing interval allocation for routing can reduce simulator calls without weakening the standard conformal guarantee.
\end{abstract}

\section{Introduction}
Engineering simulations, e.g., computational fluid dynamics (CFD), finite element analysis (FEA), and electronic design automation (EDA), are cornerstones of modern design workflows. The cost of high-fidelity simulation varies substantially with the physical problem, numerical method, resolution, hardware, and implementation. Reported scientific and engineering workflows commonly require minutes to days per simulation, which makes repeated simulator calls in a design loop a major computational bottleneck \citep{bonnet2022airfrans,ashton2024ahmedml,ananthan2024machine, jiang2024circuitnet2,chai2023circuitnet}.

Recently, deep learning surrogates, including GNN-based MeshGraphNets \citep{pfaff2021meshgraphnets, nabian2024xmeshgraphnet}, Transformer-based neural fields \citep{adams2025geotransolver, wu2024transolver, luo2025transolverplusplus, alkin2025abupt}, and Neural Operators \citep{li2021fno,li2023gino, ranade2025domino}, have been developed to accelerate high-fidelity simulations. Once trained, these models can generally evaluate new inputs more cheaply than conventional simulators on the order of milliseconds to seconds. Avoiding even a fraction of simulator calls can yield substantial computational savings.

In scientific and engineering applications, particularly in safety-critical settings, realizing the computational savings from a neural surrogate requires knowing when the surrogate is sufficiently reliable for the downstream decision. Blindly relying on the surrogate's predictions risks missing engineering violations that the full simulator can identify. This motivates a per-input routing problem: use the surrogate when its prediction is sufficiently decisive for the downstream threshold test, and otherwise defer the input to the high-fidelity simulator. The two outcomes have asymmetric costs: simulation incurs substantial computation, whereas relying on an erroneous surrogate prediction can lead to an incorrect downstream decision.

\textbf{Contributions.} Our contributions are as follows:
(i) We formulate the surrogate-simulation handoff as threshold routing under conformal coverage and identify the flexibility of the nonconformity score as an opportunity to optimize interval allocation for routing rather than uncertainty estimation (\S\ref{sec:problem});
(ii) We propose Threshold-Aware Conformal Routing (\method), whose threshold-aware $\sigma$-shaping objective concentrates interval tightness near the decision boundary through importance weighting (\S\ref{sec:objective});
(iii) We show that, across various scientific and engineering datasets, \method{} reduces simulator deferrals by \textbf{14--75\%} relative to standard conformal prediction at the same coverage target. Against a variant without threshold-local weighting but with similar predictor accuracy, \method{} further reduces deferrals by \textbf{10--24\%} on four datasets, showing an additional benefit from threshold-local weighting (\S\ref{sec:main-results});
(iv) We further show that a deep-ensemble baseline targeting predictive uncertainty leads to more deferrals, providing empirical evidence that uncertainty estimation and routing-efficient interval allocation need not share the same objective (\S\ref{sec:main-results}).

\section{Related Work}
\label{sec:related}
Related work spans calibrated uncertainty for scientific surrogates, conformal prediction (CP), conformalized quantile regression (CQR), conformal efficiency optimization, decision-aware conformal methods, and selective prediction or model routing. Existing conformal methods provide coverage guarantees or optimize the generic set size. Decision-aware and deferral methods incorporate downstream utility or fallback decisions. To our knowledge, existing methods do not jointly learn to allocate conformal interval width according to distance from a fixed decision threshold while explicitly routing ambiguous inputs to a high-fidelity simulator. Our Threshold-Aware Conformal Routing (\method{}) combines these elements: it retains split-conformal marginal coverage, optimizes interval allocation for the downstream routing decision, and invokes simulation when the calibrated interval intersects the threshold. Table~\ref{tab:related} gives a compact comparison. (See  Appendix~\ref{app:related} for details.)

\vspace{-0.5cm}
\begin{table}[h!]
\caption{\textbf{Comparison of related approaches.} Representative methods cover subsets of the components combined by \method{}. ``Guarantee'' denotes a distribution-free coverage or risk guarantee, whose target differs across rows; ``CP efficiency'' denotes direct optimization of conformal set size or interval width; and ``abstain / route'' indicates methods whose primary problem formulation is selective prediction, deferral, or model routing.}
\label{tab:related}
\centering
\small
\setlength{\tabcolsep}{3pt}
\begin{tabular}{p{0.35\linewidth}ccccc}
\toprule
\textbf{Approach} & \shortstack{\textbf{Guarantee}} & \shortstack{\textbf{CP efficiency}\\\textbf{objective}} & \shortstack{\textbf{Decision-}\\\textbf{aware}} & \shortstack{\textbf{Threshold-}\\\textbf{local}} & \shortstack{\textbf{Abstain /}\\\textbf{route}} \\
\midrule
Standard CP / CQR\textsuperscript{a} & $\checkmark$ & - & - & - & - \\
Operator-learning CP\textsuperscript{b} & $\checkmark$ & - & - & - & - \\
Conformal efficiency optimization\textsuperscript{c} & $\checkmark$ & $\checkmark$ & - & - & - \\
Learning to defer\textsuperscript{d} & - & - & $\checkmark$ & - & $\checkmark$ \\
Conformalized selective regression\textsuperscript{e} & $\checkmark$ & - & - & - & $\checkmark$ \\
Decision-aware CP\textsuperscript{f} & $\checkmark$ & - & $\checkmark$ & - & - \\
Risk-controlled surrogate routing\textsuperscript{g} & $\checkmark$ & - & $\checkmark$ & - & $\checkmark$ \\
Surrogate verification via CP\textsuperscript{h} & $\checkmark$ & - & - & - & - \\
\textbf{\method{} (ours)} & $\checkmark$ & $\checkmark$ & $\checkmark$ & $\checkmark$ & $\checkmark$ \\
\midrule
\multicolumn{6}{p{0.96\linewidth}}{\footnotesize \textsuperscript{a}~\citep{vovk2005algorithmic,romano2019cqr}; \textsuperscript{b}~\citep{ma2024calibrated}; \textsuperscript{c}~\citep{stutz2022conftr,noorani2024vrconftr,lei2023doicr,xie2024boosted,su2026cocp}; \textsuperscript{d}~\citep{madras2018predict,mozannar2020consistent}; \textsuperscript{e}~\citep{sokol2024conformalized}; \blue{\textsuperscript{f}~\citep{lekeufack2024conformal,cortesgomez2025utility,kiyani2025decision,yeh2025endtoend,yeh2025conformalrisk}}; \textsuperscript{g}~\citep{uddin2026proactive}; \textsuperscript{h}~\citep{fan2020statistical}. Standard CP, conformalized selective regression, and conformal-efficiency methods provide marginal coverage; operator-learning CP provides a finite-sample guarantee on average pointwise coverage across the output domain; decision-aware methods provide coverage or decision-risk control; risk-controlled routing bounds routed-set violations; and surrogate verification provides specification-level probabilistic validity. The coverage guarantee for conformalized selective regression is not conditional on acceptance.} \\
\bottomrule
\end{tabular}
\vskip -1.6em
\end{table}

\section{Threshold-Aware Conformal Routing (\method)}
\label{sec:method}

\begin{figure}[t]
    \centering
    \vskip -0.3cm
    \includegraphics[width=\linewidth]{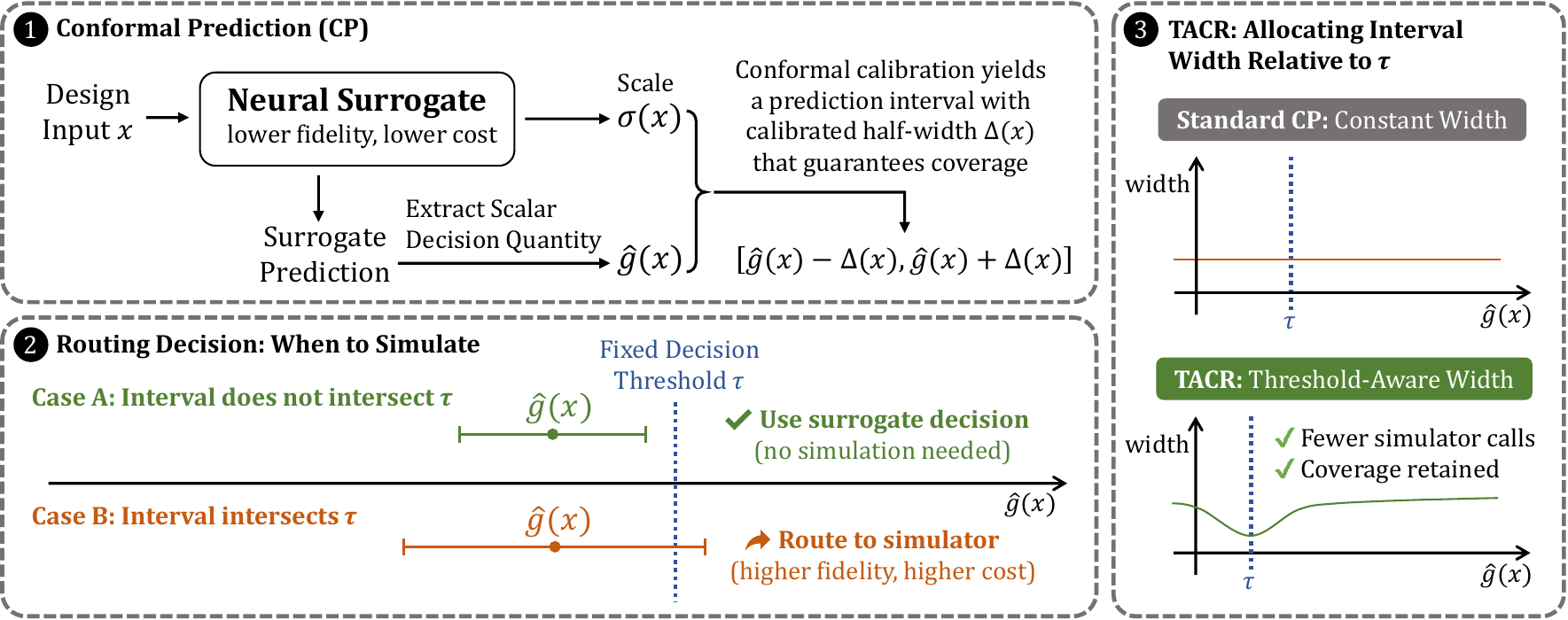}
    \vskip -0.3cm
    \caption{\textbf{Overview of Threshold-Aware Conformal Routing.}
    The calibrated interval determines whether to use the surrogate decision or route to simulation.
    \method{} allocates interval width relative to $\tau$ to reduce simulator calls while retaining conformal coverage.}
    \label{fig:overview}
    \vskip -0.3cm
\end{figure}

Figure~\ref{fig:overview} illustrates the key idea of \method{}: conformal intervals determine whether to use the surrogate or route to simulation, while interval width is allocated relative to the decision threshold $\tau$ to reduce simulator calls. We next formalize the routing objective, training objective, and split-conformal calibration procedure.

\subsection{Problem Formulation and Routing Objective}
\label{sec:problem}

A broad class of downstream engineering decisions can be expressed as threshold tests on scalar-valued functionals of high-dimensional fields produced by simulations. These functionals may take the form of integrals, extrema, percentiles, or other task-specific summaries of the full field. This enables decisions such as whether drag or peak stress exceeds a prescribed limit. We focus on this class of specification-driven decisions. Let $x$ denote the simulator input, $y$ the high-fidelity output, and $\yh=f_\theta(x)$ the surrogate prediction. A functional $g(\cdot)$ maps each output to a scalar decision quantity, giving the true value $g(y)$ and surrogate estimate $\gh(x)=g(f_\theta(x))$. Both $g$ and the decision threshold $\tau$ are specified before training. For notational convenience, define the true and surrogate threshold decisions as $d(y)=\mathbb{1}[g(y)\leq\tau]$ and $\hat d(x)=\mathbb{1}[\gh(x)\leq\tau]$, respectively. \blue{Our aim is to optimize the downstream threshold decision: for field-valued outputs, $g$ extracts the application-relevant property, e.g., a percentile, lift coefficient, or drag coefficient (Table~\ref{tab:datasets}), while $\mathcal L_{\mathrm{pred}}$ retains the surrogate's original field-level prediction objective.}

We augment the surrogate with a positive, sample-dependent scale $\sigma(x)>0$. After split-conformal calibration, this scale defines the interval
\begin{equation}
\label{eq:interval}
C(x)=[\gh(x)-\qh\sigma(x),\,\gh(x)+\qh\sigma(x)],
\end{equation}
where $\qh$ is computed on held-out calibration data. If $C(x)$ lies entirely above or below $\tau$, we use the surrogate's threshold decision; if it crosses $\tau$, we route the sample to simulation. Equivalently,
\begin{equation}
\label{eq:route}
r(x)=\mathbb{1}\!\left[|\gh(x)-\tau|\leq\qh\sigma(x)\right],
\end{equation}
where $r(x)=1$ denotes routing to simulation.

The routing rule links conformal coverage to decision safety. Whenever $r(x)=0$ and $g(y)\in C(x)$, the entire interval and the true decision quantity lie on the same side of $\tau$, so $\hat d(x)=d(y)$.  Consequently, a wrong decision that is not routed can occur only when the conformal interval fails to cover $g(y)$ (see Section~\ref{sec:calibration} for formalization of the resulting risk bound).

Our goal is to reduce the expected routing rate while retaining split-conformal coverage at a prescribed level $1-\alpha$:
\begin{equation}
\label{eq:routing-objective}
\min_{\theta,\phi}\;\mathbb{E}[r(X)]
\qquad
\text{subject to}
\qquad
\Pr\!\left(g(Y)\in C(X)\right)\geq 1-\alpha,
\end{equation}
where $\phi$ parameterizes $\sigma$ (\S\ref{sec:objective}). The coverage constraint is enforced after training by standard split-conformal calibration, rather than by treating $\alpha$ as a direct budget on decision errors. \blue{Equation~\ref{eq:route} makes the source of routing explicit. Let $h(x)=\qh\sigma(x)$ denote the calibrated half-width and $m(x)=|\gh(x)-\tau|$ the predicted threshold margin. An input is routed exactly when $h(x)\geq m(x)$; equivalently, the surrogate is used when the calibrated half-width is smaller than the predicted distance to the threshold. Consequently, inputs whose predictions lie closer to $\tau$ require correspondingly narrower intervals to avoid simulation. This motivates reducing calibrated interval width globally while applying additional width pressure near the decision threshold. Because the routing indicator is discrete and $\qh$ is available only after calibration, we optimize a differentiable training-time proxy based on the soft quantile introduced below.}

\subsection{Threshold-Aware Training Objective}
\label{sec:objective}
\textbf{Objective overview.}
Let $\mathcal B$ denote a training mini-batch. For each $i\in\mathcal B$, we define
\begin{equation}
\gh_i=g(f_\theta(x_i)),
\qquad
\sigma_i=\sigma(x_i).
\end{equation}
The complete training objective and its two scale-learning terms are
\begin{align}
\mathcal L
&=\mathcal L_{\mathrm{pred}}+\lambda\left(\mathcal L_{\mathrm{route}}+\mu\mathcal L_{\mathrm{scale}}\right),
\label{eq:loss}\\
\mathcal L_{\mathrm{route}}
&=\tilde q_{\mathcal B}\frac{1}{|\mathcal B|}\sum_{i\in\mathcal B}\left(1+\beta\prox_i\right)\sigma_i,
\\
\mathcal L_{\mathrm{scale}}
&=\frac{1}{|\mathcal B|}\sum_{i\in\mathcal B}\left(\sigma_i+\frac{1}{\sigma_i}\right).
\end{align}
Here, $\mathcal L_{\mathrm{pred}}$ denotes the surrogate's prediction objective, which may vary across architectures and datasets. The quantity $\tilde q_{\mathcal B}$ approximates the conformal multiplier during training, and $\prox_i$ measures the proximity of sample $i$ to the decision threshold. The coefficient $\lambda\geq0$ weights the two scale-learning terms relative to prediction, $\mu\geq0$ controls scale regularization, and $\beta\geq0$ controls the additional emphasis near the threshold. We define $\tilde q_{\mathcal B}$ and $\prox_i$ next.

\textbf{Differentiable conformal width.}
For each sample, define the normalized score
\begin{equation}
s_i=|g(y_i)-\gh_i|/\sigma_i.
\end{equation}
The final conformal multiplier $\qh$ is unavailable during training because it is computed later on held-out calibration data. We construct a differentiable mini-batch approximation from the normalized scores $\{s_i\}_{i\in\mathcal B}$:
\begin{equation}
\tilde q_{\mathcal B}=\operatorname{SoftQuantile}_{1-\alpha}\!\left(\{s_i\}_{i\in\mathcal B}\right).
\end{equation}
This approximation is computed separately for each training mini-batch and is used only for optimization; implementation details are given in Appendix~\ref{app:impl}. Including $\tilde q_{\mathcal B}$ couples the learned scales to the calibration mechanism, making $\tilde q_{\mathcal B}\sigma_i$ a training-time approximation to the calibrated half-width. The objective can shape the relative allocation of calibrated width rather than shrink raw scales without accounting for recalibration. 
\blue{Under a uniform rescaling $\sigma_i\mapsto c\sigma_i$, the normalized-score quantile transforms inversely as $\tilde q_{\mathcal B}\mapsto\tilde q_{\mathcal B}/c$, leaving the approximate calibrated half-width unchanged. The scale regularizer $\mathcal L_{\mathrm{scale}}$ breaks the resulting global scale indeterminacy and discourages extreme scales.}

\textbf{Threshold-local weighting.}
We define
\begin{equation}
\label{eq:proximity}
\prox_i=\exp\!\left(-\frac{\left(\operatorname{stopgrad}(\gh_i)-\tau\right)^2}{2\,\mathrm{bw}^2}\right),
\end{equation}
where $\mathrm{bw}>0$ controls the neighborhood around the threshold and is fixed during training (see Section~\ref{sec:setup} for the setting used in our experiments). The weight approaches $1$ near $\tau$ and decays toward $0$ with distance. With $\beta\geq0$, the factor $1+\beta\prox_i$ applies baseline width pressure to every sample and additional pressure near the decision boundary, where calibrated width is most likely to change the routing decision. \blue{Together with the batch-wide multiplier $\tilde q_{\mathcal B}$, these baseline and local terms turn the width–margin intuition above into a differentiable, globally coupled objective for interval allocation. See Appendix~\ref{app:direct-objective} for a comparison of this structured weighted-width surrogate with a direct smooth relaxation of the routing indicator.}

\textbf{Stop-gradient in the proximity weight.}
The dependence of $\prox_i$ on $\gh_i$ can create a shortcut when differentiated end to end. Because $\prox_i$ decreases as $|\gh_i-\tau|$ grows, the predictor can reduce the additional weight assigned to a sample by pushing its decision value away from the threshold, even when doing so does not improve prediction accuracy. We stop gradients through $\gh_i$ when computing $\prox_i$, which blocks this degenerate shortcut.

\textbf{Scale-head instantiation.}
\label{sec:sigmahead}
We parameterize the positive scale as
$\sigma(x)=\sigma_\phi(h_\theta(x))$,
where $h_\theta(x)$ is the surrogate's penultimate backbone representation and $\phi$ denotes the scale-head parameters. The lightweight head reuses backbone features and outputs one positive scale per sample. See Appendix~\ref{app:impl} for details on the architectural specifications.

\textbf{Training schedule.}
The scale head and threshold-aware objective build on the predictor's outputs and backbone features. We train in two stages: first optimizing the predictor with $\mathcal L_{\mathrm{pred}}$, then activating the scale head and jointly optimizing Eq.~\ref{eq:loss}. We use this schedule for all datasets, including CircuitNet, whose ResNet18 encoder is initialized from ImageNet weights. See Appendix~\ref{app:impl} for the stage lengths and optimization settings.

\subsection{Calibration and Deployment}
\label{sec:calibration}

After training, we freeze the predictor and scale function and perform standard split-conformal calibration on a held-out set indexed by $\mathcal I_{\mathrm{cal}}$, with $n_{\mathrm{cal}}=|\mathcal I_{\mathrm{cal}}|$. For every calibration sample, we compute
\begin{equation}
s_i=|g(y_i)-\gh(x_i)|/\sigma(x_i),
\qquad i\in\mathcal I_{\mathrm{cal}}.
\end{equation}
Let $s_{(1)}\leq\cdots\leq s_{(n_{\mathrm{cal}})}$ denote the sorted calibration scores. The finite-sample corrected conformal multiplier is
\begin{equation}
\label{eq:conformal-quantile}
k=\left\lceil(n_{\mathrm{cal}}+1)(1-\alpha)\right\rceil,
\qquad
\qh=s_{(k)},
\end{equation}
with the standard convention $\qh=\infty$ if $k>n_{\mathrm{cal}}$. At deployment, we form $C(x)$ using Eq.~\ref{eq:interval} and apply the routing rule in Eq.~\ref{eq:route}.

Under the standard exchangeability assumptions of split conformal prediction, the resulting interval satisfies distribution-free marginal coverage,
\begin{equation}
\Pr\!\left(g(Y)\in C(X)\right)\geq1-\alpha.
\end{equation}
Since a wrong threshold decision that is not routed implies noncoverage,
\begin{equation}
\label{eq:miss-bound}
\begin{aligned}
\Pr\!\left(\hat d(X)\neq d(Y),\,r(X)=0\right)
&=\Pr\!\left(\hat d(X)\neq d(Y),\,r(X)=0,\,g(Y)\notin C(X)\right)\\
&\leq\Pr\!\left(g(Y)\notin C(X)\right)\leq\alpha.
\end{aligned}
\end{equation}
The first inequality can be strict: a value outside the interval may still lie on the same side of $\tau$ as the surrogate decision. Thus, $\alpha$ specifies an interval-coverage level and provides a conservative upper bound on the unconditional miss probability; it is not a miss budget that the method is expected to exhaust. Importantly, the guarantee does not require $\sigma$ itself to estimate predictive uncertainty.

\section{Empirical Evaluation}
\label{sec:results}
We first describe the datasets, baselines, metrics, and evaluation protocol. We then compare routing performance and study sensitivity to the decision threshold, backbone, and threshold-local weighting strength.

\subsection{Experimental Design}
\label{sec:setup}

\textbf{Datasets.}
We evaluate \method{} on six datasets spanning EDA, fluid mechanics, and structural mechanics, with convolutional, graph, transformer, and MLP surrogates: CircuitNet \citep{jiang2024circuitnet2}, Darcy Flow \citep{takamoto2022pdebench}, AirfRANS \citep{bonnet2022airfrans}, AhmedML \citep{ashton2024ahmedml}, ShapeNet-Car \citep{umetani2018learning}, and a synthetic parametric-beam dataset (Table~\ref{tab:datasets}). For field-valued tasks, $g$ extracts the scalar decision quantity from the predicted field; for Beams, the surrogate predicts log-transformed peak stress directly. This diversity tests whether \method{} generalizes across backbones and output representations.

\begin{table}[t]
\vskip -1em
\caption{\textbf{Datasets and decision tasks.}
$N$ is the number of samples; p95 denotes the 95th percentile. IC, EDA, and CFD denote integrated circuit, electronic design automation, and computational fluid dynamics; $C_L$ and $C_D$ are lift and drag coefficients.}
\vskip -0.4cm
\label{tab:datasets}
\begin{center}
\small
\setlength{\tabcolsep}{4pt}
\resizebox{\linewidth}{!}{%
\begin{tabular}{llrll}
\toprule
\textbf{Dataset} & \textbf{Physics} & \textbf{$N$} & \textbf{Input $\to$ target $y$} & \textbf{Decision $g(y)$} \\
\midrule
CircuitNet & IC congestion (EDA) & 3,456 & feature map (3 channels) $\to$ congestion map & p95($y$) \\
Darcy Flow & 2D steady-state flow & 10,000 & coefficient field $128^2$ $\to$ pressure field $u$ & p95($|u|$) \\
AirfRANS & 2D airfoil CFD & 1,000 & point cloud (7 features) $\to$ surface pressure & $C_L$ \\
AhmedML & 3D bluff-body CFD & 498 & point cloud (6 features) $\to$ surface pressure & $C_D$ \\
ShapeNet-Car & 3D vehicle CFD & 888 & point cloud (xyz+normals) $\to$ surface pressure & $C_D$ \\
Beams & parametric beam (synthetic) & 2,000 & design vector $\to\log_{10}$ peak stress & $g(y)=y$ \\
\bottomrule
\end{tabular}
}
\end{center}
\vskip -0.4cm
\end{table}

\textbf{Baselines.}
Within each dataset, all methods use the same predictor backbone, with method-specific training objectives and output heads where required: \textbf{Standard conformal prediction (Standard CP)} uses a constant scale, $\sigma{=}1$; \textbf{Conformalized Quantile Regression (CQR)} \citep{romano2019cqr} uses conditional-quantile heads with $\sigma=(q_{hi}-q_{lo})/2$; \textbf{Threshold-Weighted Uncertainty Quantification (\twuq{})} adds a threshold-weighted MSE on the scalar decision quantity and trains a scale head with Gaussian negative log likelihood; \textbf{Uniform} is identical to \method{} except that it sets $\beta{=}0$, removing threshold-local weighting; \textbf{Deep Ensemble} \citep{lakshminarayanan2017deepensembles} uses the predictive standard deviation of five predictor members as $\sigma$; and our \textbf{\method} uses Eq.~\ref{eq:loss} with $\beta{=}16$ by default. Together, these baselines distinguish the effects of learned input-dependent scaling, threshold-local weighting beyond the Uniform objective, and uncertainty-oriented scaling for routing efficiency. See Appendix~\ref{app:impl} for the full optimization settings.

\textbf{Metrics.}
We report the simulation rate (\Sim{}), the percentage of test inputs routed to simulation, corresponding to the event $r(X)=1$; the miss rate (\Miss{}), the percentage for which the surrogate decision is incorrect and the input is not routed, corresponding to $\hat d(X)\neq d(Y),r(X)=0$; and $g$-MAE, the test-set mean of $|\gh(x)-g(y)|$, which measures error in the scalar decision quantity. \Sim{} is our primary efficiency metric because it directly measures the fraction of inputs that incur a high-fidelity simulator call and serves as a direct proxy for simulation cost, while \Miss{} quantifies the accompanying decision risk. For additional routing diagnostics, we define \textsc{Needs}\% as the method-specific percentage of test inputs satisfying $\hat d(X)\neq d(Y)$. See Appendix~\ref{app:full} for metric definitions and for routing precision, recall, and F1 metrics, where we treat $\hat d(X)\neq d(Y)$ as the positive class.

\textbf{Training and calibration.}
The data splits are fixed before model training, with dataset-specific proportions reported in Appendix~\ref{app:impl}. The calibration set is used only to compute the final conformal multiplier, while the training-time interpolated quantile is computed exclusively from training mini-batches. We set $\alpha=0.1$, corresponding to 90\% target coverage, and use the 70th percentile of the training-set decision quantity as the default threshold unless otherwise noted. To avoid dataset-specific tuning, we use the same global heuristic settings for the routing-loss coefficients and threshold bandwidth across all datasets and seeds. For \method{}, all datasets use predictor-only training followed by joint optimization of the predictor and scale head. See Appendix~\ref{app:impl} for the exact split sizes, baseline objectives, hyperparameters, and optimization settings. \blue{Appendix Table~\ref{tab:qpool} reports sensitivity to the training-time mini-batch quantile approximation.}

\textbf{Reporting protocol.}
We evaluate each non-ensemble method over five random seeds, and report the mean and standard deviation. The diagnostic controls use the seed counts stated in their captions. Deep Ensemble uses five independently initialized predictors per dataset and is substantially more expensive than the single-model methods because all five predictors must be trained and evaluated \citep{mouli2024using}. We report point estimates rather than seed variation since the results comprise one five-member ensemble run per dataset.

\subsection{Main Results}
\label{sec:main-results}

Table~\ref{tab:main} shows the results at the common nominal coverage level $1-\alpha=90\%$ using a fixed decision threshold for each dataset. We prioritize \Sim{} because every routed input incurs a high-fidelity simulator call, which makes it a direct proxy for simulation cost. \method{} lowers the mean \Sim{} relative to Standard CP on all six datasets and attains the lowest mean on four. \Miss{} provides the corresponding decision-risk measure and is theoretically upper-bounded by the noncoverage rate. For \method{}, it remains below 1\% on five datasets and is 2.38\% on AhmedML, which is well below the nominal 10\% noncoverage level. The $g$-MAE diagnostic measures the accuracy of the scalar decision quantity, which can affect routing efficiency by changing how accurately the surrogate resolves the threshold decision. \method{} has the lowest or second-lowest mean $g$-MAE on five of six datasets, with Beams as the exception. We distinguish below between the overall gains relative to Standard CP and the contribution of threshold-local interval allocation. \blue{See  Appendix~\ref{app:matched-budget} for a matched-budget comparison that evaluates Standard CP and CQR under the same total $E_1+E_2$ training budget; the routing gains over Standard CP persist under this control. See Appendix~\ref{sec:frontier-aligned} for a comparison of the post-hoc cost--risk trade-offs under the aligned control settings. See Appendix~\ref{app:matched-budget} for the complete matched-budget decomposition into simulation rate, false positives, the surrogate-specific zero-miss reference, and misses.}

\begin{table}[t]
\vskip -1em
\caption{\textbf{Main routing results across six scientific and engineering datasets.} \Sim{}, the fraction of inputs sent to high-fidelity simulation, is our primary cost metric; \Miss{} measures incorrect surrogate decisions that escape simulation. At the common 90\% coverage target, \method{} reduces \Sim{} relative to Standard CP on all six datasets, attains the lowest mean \Sim{} on four, and keeps \Miss{} below 2.5\% throughout. The decision-quantity mean absolute error, $g$-MAE, is an auxiliary diagnostic of predictor accuracy. Entries are mean$\pm$std over five seeds; the ensemble uses one five-member run. Lower is better. Bold and underline mark the lowest and second-lowest mean within each dataset and metric, respectively; exact ties share the same formatting. On Darcy, $\dagger$ marks CQR and \method{}, whose mean \Sim{} values differ by less than either reported standard deviation.}
\vskip -0cm
\label{tab:main}
\begin{center}
\scriptsize
\setlength{\tabcolsep}{2.5pt}
\setlength{\aboverulesep}{0.2ex}
\setlength{\belowrulesep}{0.2ex}
\begin{tabular*}{\linewidth}{@{\hspace{\tabcolsep}}cl@{\extracolsep{\fill}}cccccc@{\hspace{\tabcolsep}}}
\toprule
& Method & CircuitNet & Darcy & AirfRANS & AhmedML & ShapeNet-Car & Beams \\
\midrule
\multirow{6}{*}{\rotatebox[origin=c]{90}{\Sim{}}} & Standard CP & 46.4$\pm$0.6 & 14.8$\pm$1.1 & 19.8$\pm$2.0 & 31.7$\pm$5.1 & 46.4$\pm$3.2 & 9.4$\pm$0.7 \\
& CQR & 65.9$\pm$1.5 & \underline{12.5$\pm$5.6}$^{\dagger}$ & 19.1$\pm$3.3 & 27.9$\pm$2.4 & 38.3$\pm$3.6 & 7.8$\pm$0.6 \\
& Uniform ($\beta{=}0$) & \underline{15.3$\pm$0.8} & 14.1$\pm$0.7 & \underline{9.7$\pm$0.9} & \textbf{25.5$\pm$4.7} & \underline{36.4$\pm$3.7} & 7.5$\pm$0.6 \\
& TW-UQ & 22.5$\pm$1.0 & 13.6$\pm$1.2 & 10.5$\pm$1.0 & 36.6$\pm$1.8 & 48.2$\pm$4.8 & \underline{7.2$\pm$0.2} \\
& Deep Ensemble & 50.5 & \textbf{8.3} & 28.0 & 29.7 & 48.6 & 14.2 \\
& \textbf{\method{}} & \textbf{11.6$\pm$0.5} & \underline{12.7$\pm$0.4}$^{\dagger}$ & \textbf{7.4$\pm$0.9} & \underline{25.7$\pm$4.7} & \textbf{34.7$\pm$4.2} & \textbf{5.7$\pm$0.2} \\
\midrule
\multirow{6}{*}{\rotatebox[origin=c]{90}{\Miss{}}} & Standard CP & 0.37$\pm$0.05 & 0.03$\pm$0.02 & 0.20$\pm$0.24 & \underline{1.98$\pm$0.63} & 0.78$\pm$0.27 & \textbf{0.20$\pm$0.10} \\
& CQR & \textbf{0.09$\pm$0.09} & 0.11$\pm$0.12 & 0.30$\pm$0.40 & \textbf{1.78$\pm$1.15} & 0.67$\pm$0.22 & \textbf{0.20$\pm$0.19} \\
& Uniform ($\beta{=}0$) & 0.23$\pm$0.07 & \underline{0.02$\pm$0.02} & \underline{0.10$\pm$0.20} & 2.57$\pm$0.49 & \underline{0.56$\pm$0.35} & 0.35$\pm$0.12 \\
& TW-UQ & \underline{0.12$\pm$0.00} & 0.13$\pm$0.05 & 0.40$\pm$0.20 & \textbf{1.78$\pm$0.97} & \textbf{0.45$\pm$0.42} & 0.40$\pm$0.12 \\
& Deep Ensemble & \underline{0.12} & \textbf{0.00} & \textbf{0.00} & 2.97 & \underline{0.56} & \underline{0.25} \\
& \textbf{\method{}} & 0.44$\pm$0.11 & 0.06$\pm$0.05 & 0.40$\pm$0.20 & 2.38$\pm$0.49 & \underline{0.56$\pm$0.35} & 0.60$\pm$0.30 \\
\midrule
\multirow{6}{*}{\rotatebox[origin=c]{90}{$g$-MAE}} & Standard CP & 0.0082$\pm$0.0002 & 0.0135$\pm$0.0005 & 0.0777$\pm$0.0063 & 0.0172$\pm$0.0011 & 0.0132$\pm$0.0006 & 0.1314$\pm$0.0023 \\
& CQR & 0.0127$\pm$0.0005 & 0.0131$\pm$0.0004 & 0.0678$\pm$0.0141 & 0.0153$\pm$0.0017 & 0.0110$\pm$0.0008 & \underline{0.1302$\pm$0.0026} \\
& Uniform ($\beta{=}0$) & \textbf{0.0023$\pm$0.0001} & \underline{0.0129$\pm$0.0003} & \textbf{0.0368$\pm$0.0026} & \underline{0.0149$\pm$0.0008} & \underline{0.0101$\pm$0.0009} & 0.1320$\pm$0.0025 \\
& TW-UQ & 0.0053$\pm$0.0002 & 0.0152$\pm$0.0005 & 0.0485$\pm$0.0010 & 0.0208$\pm$0.0032 & 0.0140$\pm$0.0006 & 0.1329$\pm$0.0017 \\
& Deep Ensemble & 0.0083 & \textbf{0.0127} & 0.0694 & 0.0190 & 0.0129 & \textbf{0.1286} \\
& \textbf{\method{}} & \underline{0.0024$\pm$0.0001} & \underline{0.0129$\pm$0.0003} & \underline{0.0369$\pm$0.0032} & \textbf{0.0148$\pm$0.0007} & \textbf{0.0095$\pm$0.0008} & 0.1336$\pm$0.0024 \\
\bottomrule
\end{tabular*}
\end{center}
\vskip -0.7cm
\end{table}

\begin{wrapfigure}{r}{0.47\linewidth}
\vskip -0.6cm
\centering
\includegraphics[width=\linewidth]{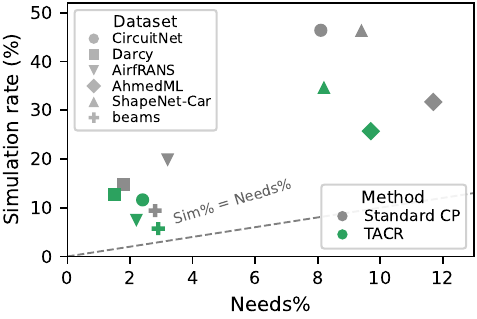}
\vskip -0.3cm
\caption{\textbf{Routing relative to the zero-miss reference.} Compared with Standard CP, \method{} requires less excess routing on all six datasets. The dashed line, $\textsc{Sim}\%=\textsc{Needs}\%$, marks the minimum routing rate under zero-miss oracle routing for each fixed surrogate.}
\label{fig:needs}
\vskip -0.2cm
\end{wrapfigure}

\textbf{Threshold-local weighting contributes to the routing gains.}
Table~\ref{tab:main} provides our primary evidence that threshold-local weighting contributes beyond changes in predictor accuracy. Relative to Uniform, \method{} reduces \Sim{} by 10--24\% on CircuitNet, Darcy, AirfRANS, and Beams. The differences on AhmedML and ShapeNet-Car are smaller than the reported standard deviations for both methods. Uniform retains \method{}'s decision-level training signal but removes threshold-local weighting, while attaining similar $g$-MAE and \textsc{Needs}\% (see Table~\ref{tab:full} in Appendix~\ref{app:full}). Because the two methods have similar predictor accuracy, their routing difference on the first four datasets is unlikely to result from predictor quality alone. In contrast, the comparison with Standard CP reflects both changes in predictor accuracy and interval allocation. An additional control in Appendix~\ref{app:g-supervision} shows that adding explicit $g$-level supervision to the baselines does not reproduce the routing gains on CircuitNet and AirfRANS. Together, these comparisons identify threshold-local interval allocation as a contributor to the routing gains.
\blue{A frozen-predictor control holds predictor quality fixed and finds lower mean \Sim{} for \method{} than Uniform on all six datasets, with similar observed \Miss{} (see Appendix~\ref{app:frozen-predictor}).}

\textbf{Predictive-uncertainty scales do not reduce simulator use.}
CQR, \twuq{}, and Deep Ensemble adapt interval width using conditional quantiles, a likelihood-based scale, and ensemble dispersion, respectively. These constructions are designed to reflect predictive uncertainty, but they do not explicitly optimize interval allocation around the decision threshold $\tau$. Table~\ref{tab:main} shows that at the common coverage target, \twuq{} routes more inputs than \method{} on all six datasets, and CQR and Deep Ensemble do so on five of the six. Uncertainty-aware scaling does not necessarily reduce \Sim{}. \blue{In addition to these uncertainty-oriented baselines, we report complementary comparisons with a direct smooth routing objective (Appendix~\ref{app:direct-objective}), Conformal Risk Training (Appendix~\ref{app:crt}), and an Uddin-inspired calibrated gate (Appendix~\ref{app:uddin}).}

\textbf{\method{} has less excess routing above the zero-miss bound.}
For a fixed surrogate, the oracle zero-miss router routes exactly those inputs on which the surrogate makes an incorrect threshold decision. This yields $\textsc{Sim}\%=\textsc{Needs}\%$, which is the minimum simulation rate achievable with zero misses. Figure~\ref{fig:needs} shows that \method{} has less excess routing above this bound than Standard CP on all six datasets; Figure~\ref{fig:needs-full} in Appendix~\ref{app:full} includes all evaluated methods. Since the evaluated methods allow nonzero \Miss{}, this bound serves as a diagnostic reference rather than an equal-safety operating point.

\subsection{Robustness and sensitivity analyses}
\label{sec:robustness}
We examine whether the routing gains persist across changes in the decision threshold, threshold-local weighting strength, and surrogate backbone.

\textbf{\method{} remains effective across decision thresholds.}
The threshold $\tau$ is an application specification fixed before deployment. Table~\ref{tab:tau} varies it over the 50th, 70th, 90th, and 95th percentiles of the training-set decision quantity on CircuitNet and AirfRANS. \method{} attains the lowest mean \Sim{} in seven of eight settings and ties Uniform in the remaining one, showing that its routing gains are not specific to the p70 threshold used in the main experiments.

\begin{table}[t]
\setlength{\aboverulesep}{0.25ex}
\setlength{\belowrulesep}{0.35ex}
\caption{\textbf{Sensitivity to the decision threshold.} Threshold-local weighting remains effective as the threshold changes: \method{} attains the lowest mean \Sim{} in seven of eight settings and ties Uniform in the remaining setting. Thresholds are percentiles of the training-set decision quantity. Reported results are mean$\pm$std over five seeds, and the main experiments use p70.}
\label{tab:tau}
\vskip -0.25cm
\begin{minipage}[t]{0.49\linewidth}
\vspace{0pt}
\centering
\small
\emph{\strut CircuitNet}\par\smallskip\vskip -0.1cm
\setlength{\tabcolsep}{1pt}
\begin{tabular*}{\linewidth}{@{\hspace{\tabcolsep}\extracolsep{\fill}}lcccc@{\hspace{\tabcolsep}}}
\toprule
\textbf{Method} & \textbf{p50} & \textbf{p70} & \textbf{p90} & \textbf{p95} \\
\midrule
Standard CP & 47.5$\pm$0.8 & 46.4$\pm$0.6 & 16.4$\pm$0.4 & 5.3$\pm$0.5 \\
CQR & 69.4$\pm$5.0 & 65.9$\pm$1.5 & 31.1$\pm$4.5 & 12.7$\pm$1.7 \\
Uniform & 24.1$\pm$0.7 & 15.3$\pm$0.8 & 6.4$\pm$0.6 & \textbf{2.6$\pm$0.3} \\
\twuq{} & 25.1$\pm$1.4 & 22.5$\pm$1.0 & 10.9$\pm$0.5 & 6.0$\pm$0.2 \\
\textbf{\method} & \textbf{19.7$\pm$0.9} & \textbf{11.6$\pm$0.5} & \textbf{5.2$\pm$0.4} & \textbf{2.6$\pm$0.3} \\
\bottomrule
\end{tabular*}
\end{minipage}\hfill
\begin{minipage}[t]{0.49\linewidth}
\vspace{0pt}
\centering
\small
\emph{\strut AirfRANS}\par\smallskip\vskip -0.1cm
\setlength{\tabcolsep}{1pt}
\begin{tabular*}{\linewidth}{@{\hspace{\tabcolsep}\extracolsep{\fill}}lcccc@{\hspace{\tabcolsep}}}
\toprule
\textbf{Method} & \textbf{p50} & \textbf{p70} & \textbf{p90} & \textbf{p95} \\
\midrule
Standard CP & 18.1$\pm$1.4 & 19.8$\pm$2.0 & 15.2$\pm$0.4 & 9.4$\pm$1.5 \\
CQR & 11.7$\pm$1.9 & 19.1$\pm$3.3 & 13.7$\pm$1.4 & 7.5$\pm$0.7 \\
Uniform & 7.4$\pm$0.6 & 9.7$\pm$0.9 & 8.2$\pm$1.3 & 6.1$\pm$0.2 \\
\twuq{} & 6.5$\pm$0.5 & 10.5$\pm$1.0 & 10.5$\pm$0.8 & 6.1$\pm$0.7 \\
\textbf{\method} & \textbf{3.7$\pm$0.2} & \textbf{7.4$\pm$0.9} & \textbf{4.5$\pm$0.4} & \textbf{3.1$\pm$1.2} \\
\bottomrule
\end{tabular*}
\end{minipage}
\vskip -0.3cm
\end{table}

\textbf{Sensitivity to threshold-local weighting.}
Table~\ref{tab:beta} varies $\beta$ on CircuitNet while holding all other settings fixed. Mean \Sim{} decreases monotonically from 13.4\% at $\beta=4$ to 9.7\% at $\beta=64$. \Miss{} remains between 0.28\% and 0.44\% without a monotonic increase. This dose-response pattern provides additional evidence that stronger threshold-local weighting systematically improves routing efficiency, rather than the gain at the default $\beta=16$ arising from a single favorable operating point. We retain the prespecified $\beta=16$ used throughout the main experiments. The sweep is intended to characterize sensitivity rather than to tune $\beta$ for individual datasets.

\textbf{Routing gains are consistent across two surrogate backbones.}
Table~\ref{tab:backbone} compares GraphSAGE with a four-head Graph Transformer on AirfRANS. The method ordering is preserved, with \method{} attaining the lowest mean \Sim{} under both backbones. This provides evidence that the observed routing gains are not specific to the GraphSAGE backbone used in the main AirfRANS experiments.

\begin{table}[t]
\vskip -0em
\setlength{\aboverulesep}{0.25ex}
\setlength{\belowrulesep}{0.35ex}
\begin{minipage}[t]{0.42\linewidth}
\centering
\caption{\textbf{Sensitivity to threshold-local weighting.} Stronger weighting monotonically reduces mean \Sim{} on CircuitNet from 13.4\% to 9.7\%, while \Miss{} shows no monotonic increase. Results are mean$\pm$std over five seeds; the main experiments use $\beta=16$, and bold marks the lowest mean in each metric.}
\label{tab:beta}
\vskip 0.1cm
\small
\renewcommand{\arraystretch}{0.92}
\setlength{\tabcolsep}{2pt}
\begin{tabular*}{\linewidth}{@{\hspace{\tabcolsep}\extracolsep{\fill}}ccc@{\hspace{\tabcolsep}}}
\toprule
$\boldsymbol{\beta}$ & \textbf{\Sim{}} & \textbf{\Miss{}} \\
\midrule
4  & 13.4$\pm$0.7 & \textbf{0.28$\pm$0.17} \\
8  & 12.5$\pm$0.3 & 0.37$\pm$0.15 \\
16 & 11.6$\pm$0.5 & 0.44$\pm$0.11 \\
32 & 11.2$\pm$0.4 & 0.39$\pm$0.16 \\
64 & \textbf{9.7$\pm$0.8} & 0.39$\pm$0.17 \\
\bottomrule
\end{tabular*}
\end{minipage}\hfill
\begin{minipage}[t]{0.56\linewidth}
\centering
\caption{\textbf{Sensitivity to the surrogate backbone.} \method{} retains the lowest mean \Sim{} when the AirfRANS backbone changes from GraphSAGE to a four-head Graph Transformer, providing evidence of transfer across these architectures. \Sim{} is mean$\pm$std over five seeds, and \Miss{} is a percentage.}
\label{tab:backbone}
\vskip 0.1cm
\small
\renewcommand{\arraystretch}{0.92}
\setlength{\tabcolsep}{2pt}
\begin{tabular*}{\linewidth}{@{\hspace{\tabcolsep}\extracolsep{\fill}}llcc@{\hspace{\tabcolsep}}}
\toprule
\textbf{Backbone} & \textbf{Method} & \textbf{\Sim{}} & \textbf{\Miss{}} \\
\midrule
\multirow{4}{*}{GraphSAGE} & Standard CP & 19.8$\pm$2.0 & 0.20 \\
& CQR & 19.1$\pm$3.3 & 0.30 \\
& Uniform ($\beta{=}0$) & 9.7$\pm$0.9 & 0.10 \\
& \textbf{\method} & \textbf{7.4$\pm$0.9} & 0.40 \\
\midrule
\multirow{4}{*}{\shortstack[l]{Graph-\\Transformer}} & Standard CP & 18.3$\pm$1.7 & 0.10 \\
& CQR & 15.2$\pm$0.9 & 0.10 \\
& Uniform ($\beta{=}0$) & 8.6$\pm$1.4 & 0.10 \\
& \textbf{\method} & \textbf{6.3$\pm$0.5} & 0.20 \\
\bottomrule
\end{tabular*}
\end{minipage}
\vskip -0.3cm
\end{table}

\subsection{Post-hoc cost-risk trade-offs}
\label{sec:frontier}
\begin{figure}[t]
\centering
\includegraphics[width=\linewidth]{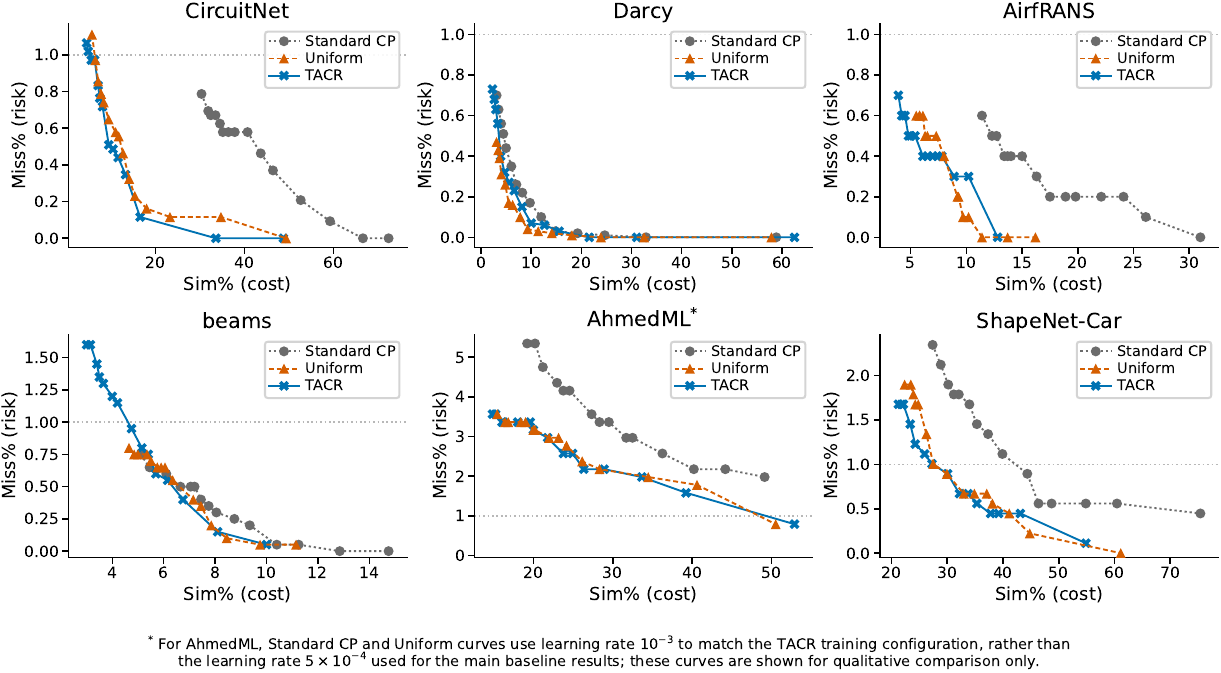}
\vskip -0.2cm
\caption{\textbf{Post-hoc cost--risk trade-offs.}
Models are trained once with $\alpha=0.1$ and then recalibrated over $\alpha\in[0.02,0.30]$ by recomputing only the exact conformal quantile, without retraining. Trade-offs vary by dataset and operating point: learned-scale methods often require fewer simulator calls than Standard CP at comparable \Miss{}, while crossings between \method{} and Uniform show that neither uniformly dominates across all recalibrated operating points. Curves aggregate five seeds; the lower left is preferred. The dotted horizontal line marks a 1\% miss rate. For AhmedML, Standard CP and Uniform use a learning rate of $10^{-3}$ rather than the $5\times10^{-4}$ used in the main table.}

\label{fig:frontier}
\end{figure}

The main results use models trained with $\alpha=0.1$. We next examine whether a model trained at this single coverage level can still support useful deployment-time trade-offs through recalibration alone. Without retraining, we recompute the exact conformal quantile $\qh(\alpha)$ over $\alpha\in[0.02,0.30]$ and evaluate the resulting \Sim{}--\Miss{} operating points. Figure~\ref{fig:frontier} shows that changing the calibration level produces substantial variation in simulator use and decision risk. This indicates that a fixed trained model can be adapted post hoc to different deployment requirements. Smaller $\alpha$ generally produces wider intervals and more routing, while empirical \Miss{} may decrease. Since conformal ranks and test-set counts are discrete, the curves are stepwise and empirical \Miss{} need not vary monotonically. The relative ordering is dataset- and operating-point-dependent, with crossings between \method{} and Uniform in some regimes. Since $\alpha$ also enters the training-time soft quantile, these curves characterize the post-hoc flexibility of models trained at $\alpha=0.1$, rather than separately optimized performance at each coverage level.

\nopagebreak

\section{Conclusion}

Learned surrogates can reduce the cost of high-fidelity simulation, but their errors can alter downstream decisions. We formulate the surrogate-simulator handoff for fixed scalar thresholds as threshold-aware routing with split-conformal coverage. Exploiting conformal validity for any nonconformity score fixed before calibration, our proposed \method{} method learns to allocate interval width according to its effect on routing. Across various scientific and engineering datasets, \method{} reduces the fraction of inputs routed to simulation by 14--75\% relative to standard conformal prediction at the same nominal coverage target. Against a variant without threshold-local weighting but with similar predictor accuracy, \method{} further reduces deferrals by 10--24\% on four datasets. These results show that routing-aware interval allocation can reduce simulator calls without weakening the standard conformal guarantee.

\textbf{Limitations and future work.} \method{} currently addresses decisions defined by a fixed scalar threshold. Extending it to multiple simultaneous specifications would require a joint routing rule and an appropriate calibration target. In addition, its guarantee is marginal rather than conditional, so the stratified results in Appendix~\ref{sec:conditional} are empirical diagnostics and do not establish local risk control. Routing efficiency depends on surrogate accuracy and on the distribution of decision quantities around the threshold, and may vary across tasks. Extensions to group-conditional calibration, multi-constraint decisions, and online updating with newly simulated samples are promising future directions.

\subsection*{Reproducibility statement}
Sections~\ref{sec:method} and~\ref{sec:setup} specify the method, calibration and routing procedures, datasets, baselines, metrics, and evaluation protocol; Section~\ref{sec:calibration} states the assumptions underlying the finite-sample guarantee. Appendix~\ref{app:impl} provides the exact data splits and preprocessing, dataset-specific architectures and scale heads, prediction objectives, hyperparameters, optimization schedules, and calibration details. Dataset and backbone sources are cited in Section~\ref{sec:setup} and Table~\ref{tab:datasets}, and the captions identify the seed aggregation and evaluation settings used for the reported results.

\subsection*{AI use statement}
In this work, we used generative AI tools to retrieve and organize relevant literature; assist with code development, code auditing, result aggregation, and figure preparation; and polish portions of the manuscript. The authors reviewed all AI-assisted text, checked cited literature against the original sources, and inspected and tested AI-assisted code and numerical outputs. Generative AI was not used to generate synthetic datasets or prove mathematical claims. We take responsibility for the final content of this work, including text, claims, code, and artifacts produced with the aid of generative AI.

\bibliography{iclr2027_conference}
\bibliographystyle{iclr2027_conference}
\newpage
\appendix

\section{Extended Related Work}
\label{app:related}
This section expands the comparison in Section~\ref{sec:related} and distinguishes the statistical targets and modeling objectives of the most closely related approaches.

\textbf{Selective prediction and model routing.}
Selective prediction allows a model to withhold its prediction when it is unreliable, optionally deferring to a fallback model or expert. Learning-to-defer methods jointly learn when to predict and when to delegate to a downstream expert, accounting for both model and expert performance in the system objective \citep{madras2018predict,mozannar2020consistent}. Conformalized selective regression instead uses conformalized interval width as an input-dependent rejection score \citep{sokol2024conformalized}. More generally, deep ensembles \citep{lakshminarayanan2017deepensembles}, MC-Dropout \citep{gal2016dropout}, and evidential regression \citep{amini2020evidential} provide uncertainty estimates that can serve as deferral scores when incorporated into a routing system. Learning-to-defer methods optimize the system-level handoff but do not construct conformal intervals, whereas conformalized selective regression and the uncertainty-quantification methods above use generic uncertainty measures for rejection. None of these approaches trains conformal interval allocation specifically around a downstream decision threshold.

Most closely related to our setting, \citet{uddin2026proactive} train an input-based gate that chooses between an interpretable surrogate and a reference model before either model is evaluated. They train the gate to distinguish safe from unsafe inputs and calibrate its routing threshold on held-out data to ensure, with probability at least \(1-\delta\), that the violation rate among inputs routed to the surrogate is at most $\alpha$. This guarantee concerns performance degradation conditional on routing, whereas our conformal interval provides marginal coverage for the simulator-derived decision quantity. Their method also does not learn how the interval width should be allocated around a downstream specification threshold.
\blue{Since this conditional risk differs from the unconditional miss event controlled by \method{}, Appendix~\ref{app:uddin} compares a Uddin-inspired gate with \method{} at matched empirical miss rates.}

\textbf{Conformal efficiency optimization.}
Split conformal prediction (CP) provides finite-sample marginal coverage under exchangeability \citep{vovk2005algorithmic,papadopoulos2002inductive}; \citet{angelopoulos2023gentle} give a modern introduction. Conformalized Quantile Regression (CQR) adapts interval width to heteroscedasticity by conformalizing learned conditional quantiles \citep{romano2019cqr}. Other methods optimize conformal efficiency more directly. ConfTr differentiates through a mini-batch approximation to conformal prediction to optimize the size and composition of classification sets \citep{stutz2022conftr}, while VR-ConfTr reduces the variance of its gradient estimates \citep{noorani2024vrconftr}. For regression, DOICR directly minimizes average interval width \citep{lei2023doicr}; Boosted Conformal Intervals uses gradient boosting to refine a predefined conformity score for shorter intervals or improved conditional coverage \citep{xie2024boosted}; and CoCP jointly learns interval centers and radii \citep{su2026cocp}. These methods optimize conformal efficiency or other interval-quality objectives, but their objectives are not localized by distance to an externally specified downstream decision threshold.

\textbf{Decision-aware conformal prediction.}
A growing line of work connects conformal methods with downstream decisions. Conformal Decision Theory calibrates decisions directly to control their realized average loss (decision risk), without first constructing prediction sets \citep{lekeufack2024conformal}. Utility-Directed CP incorporates a downstream cost function into the construction of classification prediction sets while retaining standard coverage guarantees \citep{cortesgomez2025utility}. \blue{Risk-Averse Calibration provides prediction sets and max-min decision policies for agents optimizing value at risk \citep{kiyani2025decision}.} End-to-end conformal calibration for optimization learns uncertainty sets for conditional robust optimization using the downstream decision loss \citep{yeh2025endtoend}. \blue{Conformal Risk Training differentiates through conformal risk control during training or fine-tuning to improve downstream performance under a controlled risk \citep{yeh2025conformalrisk}. We instantiate this framework for simulator routing in Appendix~\ref{app:crt}.} These methods show how conformal procedures can account for decision risk or utility, but they do not study when to invoke a high-fidelity simulator according to whether a scalar conformal interval crosses a fixed specification threshold.

\textbf{Scientific surrogates and calibrated uncertainty.}
Scientific surrogate modeling spans neural operators such as FNO \citep{li2021fno} and GINO \citep{li2023gino}, graph-based simulators such as MeshGraphNet \citep{pfaff2021meshgraphnets} and X-MeshGraphNet \citep{nabian2024xmeshgraphnet}, and transformer-based models such as Transolver \citep{wu2024transolver}, Transolver++ \citep{luo2025transolverplusplus}, GeoTransolver \citep{adams2025geotransolver}, and AB-UPT \citep{alkin2025abupt}. Complementary work studies uncertainty quantification and calibration for these surrogates. \citet{magnani2025bayesian} develop an approximate Bayesian framework for uncertainty quantification in neural operators. \citet{mouli2024using} evaluate several uncertainty-quantification methods for neural operators under out-of-domain inputs and introduce DiverseNO, a multi-head model designed to approximate the uncertainty benefits of an ensemble at lower computational cost. \citet{ma2024calibrated} provide distribution-free, finite-sample calibration for function-valued operator outputs, targeting the expected fraction of domain points covered by the uncertainty band, and \citet{fan2020statistical} use surrogate models and conformal inference to provide probabilistic guarantees for the satisfaction of formal system specifications. These methods quantify or calibrate surrogate outputs and formal properties, but they do not learn threshold-local conformal interval allocation for selective simulator invocation.

Table~\ref{tab:related} condenses these distinctions. Because the cited methods establish different statistical statements, a checkmark in the guarantee column indicates the presence of a distribution-free coverage or risk guarantee, not that the guaranteed quantities are equivalent.

\section{\blue{Matched-budget baseline comparison}}
\label{app:matched-budget}

\blue{We align the total training budget across all methods before decomposing simulator use. Standard CP and CQR receive the same total $E_1+E_2$ data-pass budget as \method{}; during stage 2, Standard CP continues optimizing its prediction loss and CQR its quantile objective. \twuq{}, Uniform, and \method{} use their common two-stage protocol, and each Deep Ensemble member receives the full predictor-training budget. All methods use the exact finite-sample conformal order statistic for final calibration. Table~\ref{tab:matched-budget} reports the complete comparison.}

\begin{table*}[h!]
\caption{\textbf{Complete matched-budget routing results and metric decomposition.} Standard CP, CQR, \twuq{}, Uniform, and \method{} report mean$\pm$std over five seeds for all four metrics. Deep Ensemble is one ensemble of five independently trained full-budget members, so its entries are point estimates. Bold marks the lowest mean \Sim{} and FP in each dataset.}
\label{tab:matched-budget}
\label{tab:routing-decomposition}
\centering
\scriptsize
\setlength{\tabcolsep}{3.5pt}
\renewcommand{\arraystretch}{0.88}
\begin{tabular}{llrrrr}
\toprule
\textbf{Dataset} & \textbf{Method} & \textbf{\Sim{} $\downarrow$} & \textbf{FP $\downarrow$} & \textbf{\textsc{Needs}\%} & \textbf{\Miss{} $\downarrow$} \\
\midrule
CircuitNet & Standard CP & 46.4$\pm$0.6 & 38.7$\pm$0.6 & 8.1$\pm$0.5 & 0.37$\pm$0.05 \\
& CQR & 65.9$\pm$2.3 & 52.0$\pm$2.1 & 14.0$\pm$0.8 & 0.12$\pm$0.07 \\
& \twuq{} & 22.5$\pm$1.0 & 18.1$\pm$0.9 & 4.5$\pm$0.4 & 0.12$\pm$0.00 \\
& Uniform & 15.3$\pm$0.8 & 13.2$\pm$0.8 & 2.4$\pm$0.3 & 0.23$\pm$0.07 \\
& Deep Ensemble & 47.3 & 39.9 & 7.4 & 0.00 \\
& \method{} & \textbf{11.6$\pm$0.3} & \textbf{9.9$\pm$0.4} & 2.2$\pm$0.1 & 0.44$\pm$0.11 \\
\midrule
AirfRANS & Standard CP & 22.7$\pm$2.0 & 18.9$\pm$2.1 & 3.9$\pm$1.0 & 0.10$\pm$0.20 \\
& CQR & 14.6$\pm$2.7 & 11.4$\pm$2.7 & 3.5$\pm$0.6 & 0.30$\pm$0.40 \\
& \twuq{} & 10.5$\pm$1.0 & 8.7$\pm$1.3 & 2.2$\pm$0.9 & 0.40$\pm$0.20 \\
& Uniform & 9.7$\pm$0.9 & 6.9$\pm$0.7 & 2.9$\pm$0.5 & 0.10$\pm$0.20 \\
& Deep Ensemble & 21.5 & 19.0 & 2.5 & 0.00 \\
& \method{} & \textbf{7.1$\pm$1.5} & \textbf{5.1$\pm$1.1} & 2.3$\pm$0.2 & 0.30$\pm$0.24 \\
\midrule
Darcy & Standard CP & 15.8$\pm$0.7 & 14.3$\pm$0.6 & 1.6$\pm$0.2 & 0.04$\pm$0.02 \\
& CQR & 11.7$\pm$3.6 & 10.5$\pm$3.5 & 1.3$\pm$0.1 & 0.10$\pm$0.07 \\
& \twuq{} & 13.6$\pm$1.2 & 11.7$\pm$1.1 & 2.1$\pm$0.2 & 0.13$\pm$0.05 \\
& Uniform & 14.1$\pm$0.7 & 12.8$\pm$0.7 & 1.4$\pm$0.2 & 0.02$\pm$0.02 \\
& Deep Ensemble & \textbf{6.1} & \textbf{5.1} & 1.5 & 0.45 \\
& \method{} & 12.3$\pm$1.4 & 10.9$\pm$1.4 & 1.4$\pm$0.1 & 0.07$\pm$0.04 \\
\midrule
Beams & Standard CP & 9.3$\pm$0.7 & 6.8$\pm$0.8 & 2.8$\pm$0.1 & 0.20$\pm$0.10 \\
& CQR & 7.8$\pm$0.6 & 5.4$\pm$0.5 & 2.6$\pm$0.3 & 0.20$\pm$0.19 \\
& \twuq{} & 7.2$\pm$0.2 & 4.7$\pm$0.2 & 2.9$\pm$0.2 & 0.40$\pm$0.12 \\
& Uniform & 7.5$\pm$0.6 & 5.1$\pm$0.6 & 2.7$\pm$0.2 & 0.35$\pm$0.12 \\
& Deep Ensemble & 13.5 & 11.0 & 2.8 & 0.25 \\
& \method{} & \textbf{5.8$\pm$0.3} & \textbf{3.6$\pm$0.3} & 2.8$\pm$0.3 & 0.60$\pm$0.30 \\
\midrule
AhmedML & Standard CP & 30.7$\pm$2.6 & 23.4$\pm$1.6 & 9.3$\pm$3.5 & 1.98$\pm$1.40 \\
& CQR & 27.9$\pm$2.4 & 20.4$\pm$1.3 & 9.3$\pm$2.8 & 1.78$\pm$1.15 \\
& \twuq{} & 37.4$\pm$2.9 & 26.5$\pm$4.3 & 13.1$\pm$3.9 & 2.18$\pm$2.10 \\
& Uniform & 25.9$\pm$4.1 & 17.8$\pm$2.3 & 9.3$\pm$1.9 & 1.19$\pm$0.97 \\
& Deep Ensemble & 34.7 & 24.8 & 9.9 & 0.00 \\
& \method{} & \textbf{24.8$\pm$4.3} & \textbf{16.8$\pm$2.4} & 9.1$\pm$1.6 & 1.19$\pm$0.97 \\
\midrule
ShapeNet-Car & Standard CP & 45.1$\pm$2.2 & 36.9$\pm$1.9 & 8.9$\pm$0.9 & 0.67$\pm$0.22 \\
& CQR & 38.3$\pm$3.6 & 30.6$\pm$3.0 & 8.4$\pm$1.6 & 0.67$\pm$0.22 \\
& \twuq{} & 48.2$\pm$4.8 & 38.0$\pm$5.0 & 10.6$\pm$0.8 & 0.45$\pm$0.42 \\
& Uniform & 37.1$\pm$2.5 & 30.4$\pm$1.7 & 7.4$\pm$1.4 & 0.67$\pm$0.42 \\
& Deep Ensemble & 49.7 & 41.9 & 8.4 & 0.56 \\
& \method{} & \textbf{35.3$\pm$2.9} & \textbf{28.8$\pm$2.5} & 7.0$\pm$1.4 & 0.56$\pm$0.50 \\
\bottomrule
\end{tabular}
\renewcommand{\arraystretch}{1}
\end{table*}

\blue{\method{} reduces mean \Sim{} relative to matched-budget Standard CP on all six datasets and attains the lowest \Sim{} and FP on five of six. Thus, its routing gains cannot be attributed simply to additional predictor training. Deep Ensemble often achieves a lower observed \Miss{}, but generally routes substantially more inputs; on AhmedML, for example, its zero miss rate is accompanied by a \Sim{} of 34.7\%, compared with 24.8\% for \method{}.}

\section{\blue{Post-hoc cost--risk trade-offs under aligned settings}}
\label{sec:frontier-aligned}

\begin{figure}[t]
\centering
\includegraphics[width=\linewidth]{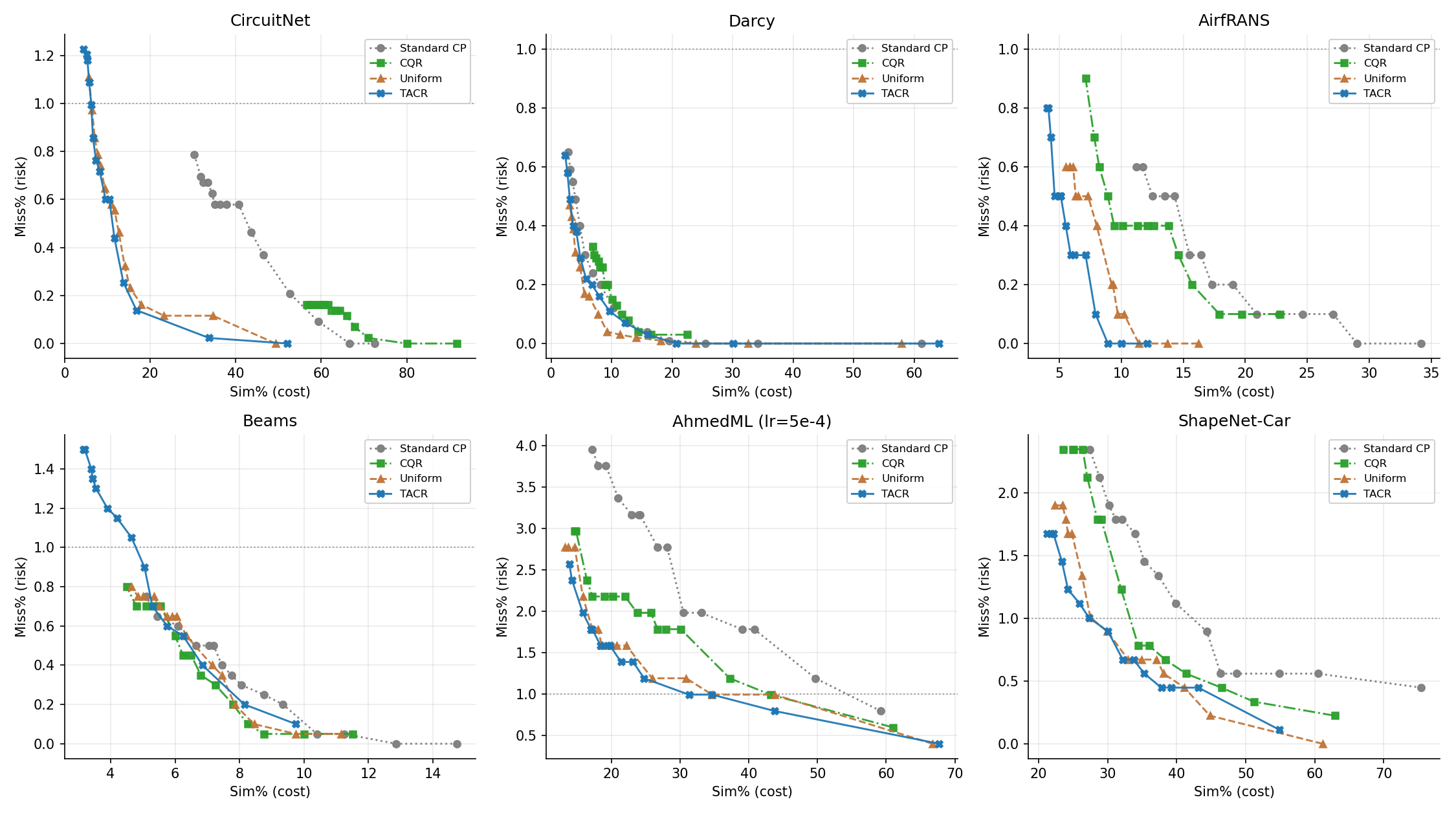}
\vskip -0.2cm
\caption{\blue{\textbf{Post-hoc cost--risk trade-offs under aligned settings.}
Models are trained once with $\alpha=0.1$ and recalibrated over $\alpha\in[0.02,0.30]$ by recomputing only the exact finite-sample conformal quantile, without retraining. The comparison includes Standard CP, CQR, Uniform, and \method{} on all six datasets. Standard CP and CQR use the matched total $E_1+E_2$ training budget on CircuitNet, AirfRANS, Darcy, and Beams, while all four AhmedML methods use the aligned base learning rate of $5\times10^{-4}$. Curves report means over five seeds; the lower left is preferred. The dotted horizontal line marks a 1\% miss rate.}}
\label{fig:frontier-aligned}
\end{figure}

\blue{We use the aligned control runs to examine whether models trained at $\alpha=0.1$ support useful deployment-time trade-offs through recalibration alone. Without retraining, we recompute the exact finite-sample conformal quantile $\qh(\alpha)$ over $\alpha\in[0.02,0.30]$ and evaluate the resulting \Sim{}--\Miss{} operating points. For CircuitNet, AirfRANS, Darcy, and Beams, Standard CP and CQR use the matched total $E_1+E_2$ training budget from Appendix~\ref{app:matched-budget}; for AhmedML, all four methods use the aligned base learning rate of $5\times10^{-4}$. Figure~\ref{fig:frontier-aligned} shows substantial variation in simulator use and decision risk across calibration levels. The curves are stepwise because conformal ranks and test-set counts are discrete, and their relative ordering depends on the dataset and operating point. Since $\alpha$ also enters the training-time soft quantile, these curves characterize the post-hoc flexibility of models trained at $\alpha=0.1$, rather than separately optimized performance at each coverage level.}

\FloatBarrier

\section{\blue{Frozen-predictor attribution}}
\label{app:frozen-predictor}

\blue{We isolate threshold-local interval allocation by holding the predictor fixed. For each dataset and seed, Standard CP, Uniform, and \method{} share the same Stage-1 predictor, whose backbone and prediction head remain frozen. Standard CP uses a constant scale, while Uniform and \method{} start from identically initialized scale heads and train only the scale path for the same $E_2$ budget; the latter two differ only through threshold-local weighting, with $\beta=0$ for Uniform and $\beta=16$ for \method{}.}

\begin{table*}[h!]
\centering
\caption{\textbf{Frozen-predictor attribution across all six datasets.} Standard CP, Uniform, and \method{} share the same frozen predictor within each dataset and seed. Entries are mean$\pm$std over five seeds. The common $g$-MAE and \textsc{Needs}\% verify that predictor accuracy is fixed; bold marks the lowest mean \Sim{} within each dataset.}
\label{tab:frozen-predictor}
\small
\setlength{\tabcolsep}{5pt}
\begin{tabular}{llrrrr}
\toprule
\textbf{Dataset} & \textbf{Method} & \textbf{Common $g$-MAE} & \textbf{Common \textsc{Needs}\%} & \textbf{\Sim{}} & \textbf{\Miss{}} \\
\midrule
\multirow{3}{*}{CircuitNet} & Standard CP & \multirow{3}{*}{$0.00821\!\pm\!0.00021$} & \multirow{3}{*}{$8.08\!\pm\!0.51$} & $46.4\!\pm\!0.6$ & $0.37\!\pm\!0.05$ \\
& Uniform & & & $44.9\!\pm\!1.1$ & $0.32\!\pm\!0.09$ \\
& \method{} & & & $\bm{43.0\!\pm\!2.8}$ & $0.19\!\pm\!0.06$ \\
\midrule
\multirow{3}{*}{AirfRANS} & Standard CP & \multirow{3}{*}{$0.07931\!\pm\!0.00973$} & \multirow{3}{*}{$3.30\!\pm\!1.75$} & $21.3\!\pm\!2.4$ & $0.30\!\pm\!0.24$ \\
& Uniform & & & $18.2\!\pm\!3.4$ & $0.10\!\pm\!0.20$ \\
& \method{} & & & $\bm{15.2\!\pm\!3.6}$ & $0.20\!\pm\!0.24$ \\
\midrule
\multirow{3}{*}{Darcy} & Standard CP & \multirow{3}{*}{$0.01310\!\pm\!0.00046$} & \multirow{3}{*}{$1.54\!\pm\!0.20$} & $14.8\!\pm\!1.2$ & $0.05\!\pm\!0.03$ \\
& Uniform & & & $15.1\!\pm\!1.3$ & $0.05\!\pm\!0.03$ \\
& \method{} & & & $\bm{12.8\!\pm\!1.0}$ & $0.08\!\pm\!0.06$ \\
\midrule
\multirow{3}{*}{Beams} & Standard CP & \multirow{3}{*}{$0.13141\!\pm\!0.00225$} & \multirow{3}{*}{$2.80\!\pm\!0.10$} & $9.3\!\pm\!0.7$ & $0.20\!\pm\!0.10$ \\
& Uniform & & & $7.5\!\pm\!0.4$ & $0.40\!\pm\!0.20$ \\
& \method{} & & & $\bm{6.1\!\pm\!0.3}$ & $0.60\!\pm\!0.30$ \\
\midrule
\multirow{3}{*}{AhmedML} & Standard CP & \multirow{3}{*}{$0.01782\!\pm\!0.00304$} & \multirow{3}{*}{$10.10\!\pm\!3.28$} & $\bm{33.1\!\pm\!3.5}$ & $1.98\!\pm\!1.40$ \\
& Uniform & & & $33.5\!\pm\!4.2$ & $1.98\!\pm\!1.40$ \\
& \method{} & & & $\bm{33.1\!\pm\!5.0}$ & $1.98\!\pm\!1.40$ \\
\midrule
\multirow{3}{*}{ShapeNet-Car} & Standard CP & \multirow{3}{*}{$0.01295\!\pm\!0.00033$} & \multirow{3}{*}{$9.16\!\pm\!1.60$} & $45.3\!\pm\!3.1$ & $1.12\!\pm\!0.35$ \\
& Uniform & & & $45.3\!\pm\!3.1$ & $1.12\!\pm\!0.35$ \\
& \method{} & & & $\bm{43.9\!\pm\!4.4}$ & $1.12\!\pm\!0.35$ \\
\bottomrule
\end{tabular}
\end{table*}

\blue{Because the predictor is shared and frozen, $g$-MAE and \textsc{Needs}\% are identical across all three arms within every seed. \method{} lowers mean \Sim{} relative to Uniform on all six datasets while maintaining similar observed \Miss{}, showing that threshold-local weighting can reduce simulator calls even when predictor quality is held fixed.}

\FloatBarrier
\section{Additional decision-level supervision diagnostic}
\label{app:g-supervision}
The comparison with Uniform in Section~\ref{sec:main-results} is our primary evidence for separating the benefit of threshold-local weighting from changes in predictor quality. As an additional check, Table~\ref{tab:fairness} augments each baseline with an extra unweighted $g$-level MSE loss while leaving \method{} without this auxiliary term. On both CircuitNet and AirfRANS, \method{} routes fewer samples than every baseline in this comparison. Thus, adding this direct supervision of the decision quantity does not by itself reproduce the routing efficiency of \method{} on these two datasets.

\begin{table}[h!]
\centering
\caption{Explicit decision-level supervision does not reproduce \method{}'s routing gains: without auxiliary $g$-MSE, \method{} attains the lowest mean \Sim{} on both datasets. Values are percentages reported as mean$\pm$std over five seeds; bold marks the lowest mean.}
\label{tab:fairness}
\small
\setlength{\tabcolsep}{6pt}
\begin{tabular}{lcc}
\toprule
\textbf{Method} & \textbf{CircuitNet} & \textbf{AirfRANS} \\
\midrule
Standard CP  & 18.5$\pm$0.6 & 11.9$\pm$0.8 \\
CQR          & 37.5$\pm$2.1 & 9.8$\pm$1.0 \\
Uniform      & 12.2$\pm$0.8 & 8.5$\pm$0.8 \\
\twuq{}      & 21.7$\pm$3.1 & 9.7$\pm$1.0 \\
\textbf{\method{} (no aux)} & \textbf{11.6$\pm$0.5} & \textbf{7.4$\pm$0.9} \\
\bottomrule
\end{tabular}
\end{table}

\section{\blue{Controlled comparison with a direct routing objective}}
\label{app:direct-objective}

\blue{Define the training-time calibrated half-width $h_i=\tilde q_{\mathcal B}\sigma_i$ and predicted margin $m_i=|\gh_i-\tau|$. The routing event is $\mathbb 1[h_i\geq m_i]$, which naturally suggests a direct smooth relaxation. The direct objective and the routing term of \method{} can be written as}
{
\begin{align}
\mathcal L_{\mathrm{direct}}
&=\frac{1}{|\mathcal B|}\sum_{i\in\mathcal B}
\operatorname{sigmoid}\!\left(\frac{h_i-m_i}{T}\right),\\
\mathcal L_{\mathrm{route}}
&=\frac{1}{|\mathcal B|}\sum_{i\in\mathcal B}
\left[1+\beta\exp\!\left(-\frac{m_i^2}{2\,\mathrm{bw}^2}\right)\right]h_i.
\end{align}
}

\blue{In both implementations, we use the detached margin $m_i=|\operatorname{stopgrad}(\gh_i)-\tau|$. This common stop-gradient is essential to the comparison. Without it, increasing $|\gh_i-\tau|$ reduces both the direct sigmoid loss and the threshold-local weight in $\mathcal L_{\mathrm{route}}$. The predictor could therefore lower either routing term by moving its decision estimate away from the threshold rather than by learning a more useful interval allocation. Detaching $m_i$ blocks this shared margin-based shortcut.}

\blue{Under this shared condition, the direct objective penalizes configurations in which $h_i$ approaches or exceeds $m_i$, while \method{} explicitly penalizes calibrated width with greater emphasis at smaller margins. Because $\tilde q_{\mathcal B}$ is determined by the batch-wide ranking of normalized residuals, scale allocation is globally coupled through the calibration multiplier. The factor $1$ retains a global average calibrated-width objective, while the exponential term adds threshold-local emphasis. This establishes a structural connection to routing, but not a population-level routing-optimality guarantee under joint training. For the controlled comparison, we use the same Stage-1 checkpoints, prediction objective, optimizer groups, loss coefficients, and joint-training schedule as \method{}, replacing only $\mathcal L_{\mathrm{route}}$ with $\mathcal L_{\mathrm{direct}}$ and setting $T=\mathrm{bw}$.}

\begin{table}[h!]
\centering
\caption{Controlled comparison with a direct smooth-routing objective. Values are percentages reported as mean$\pm$std over five seeds. Bold marks the lower mean \Sim{} for each dataset.}
\label{tab:direct-objective}
\small
\setlength{\tabcolsep}{4.5pt}
\begin{tabular}{lcccc}
\toprule
& \multicolumn{2}{c}{\textbf{\Sim{} $\downarrow$}} & \multicolumn{2}{c}{\textbf{\Miss{} $\downarrow$}} \\
\cmidrule(lr){2-3}\cmidrule(lr){4-5}
\textbf{Dataset} & \textbf{\method{}} & \textbf{Direct} & \textbf{\method{}} & \textbf{Direct} \\
\midrule
CircuitNet & \textbf{11.6$\pm$0.3} & 12.0$\pm$0.3 & 0.44$\pm$0.11 & 0.49$\pm$0.24 \\
AirfRANS   & \textbf{7.1$\pm$1.5}  & 8.3$\pm$1.5  & 0.30$\pm$0.24 & 0.10$\pm$0.20 \\
Darcy      & \textbf{12.3$\pm$1.4} & 14.1$\pm$1.3 & 0.07$\pm$0.04 & 0.04$\pm$0.04 \\
Beams      & \textbf{5.8$\pm$0.3}  & 6.5$\pm$0.4  & 0.60$\pm$0.30 & 0.50$\pm$0.22 \\
\bottomrule
\end{tabular}
\end{table}

\blue{Under otherwise identical training settings, \method{} attains a lower mean \Sim{} on all four datasets, while both objectives keep the observed \Miss{} at or below $0.60\%$. The direct relaxation is competitive, and the comparison provides empirical support for the structured weighted-width surrogate used by \method{}.}

\section{\blue{Comparison with Conformal Risk Training}}
\label{app:crt}

\blue{Conformal Risk Training (CRT) differentiates through conformal risk control to optimize a downstream objective while preserving its calibrated risk target \citep{yeh2025conformalrisk}. We instantiate CRT for simulator routing by using scalar-interval miscoverage as the controlled risk and the routing indicator as the downstream cost. CRT and \method{} use the same Stage-1 checkpoints, nominal coverage target, $E_2$ training budget, and optimizer settings. Following CRT's nested construction, risk calibration and downstream-objective evaluation use separate mini-batch partitions.}

\begin{table}[h!]
\centering
\caption{Comparison with a routing instantiation of Conformal Risk Training. Entries report percentages as mean$\pm$std over five seeds. Bold marks the lower mean \Sim{} for each dataset.}
\label{tab:crt}
\small
\setlength{\tabcolsep}{4.5pt}
\begin{tabular}{lcccc}
\toprule
& \multicolumn{2}{c}{\textbf{\Sim{} $\downarrow$}} & \multicolumn{2}{c}{\textbf{\Miss{} $\downarrow$}} \\
\cmidrule(lr){2-3}\cmidrule(lr){4-5}
\textbf{Dataset} & \textbf{\method{}} & \textbf{CRT} & \textbf{\method{}} & \textbf{CRT} \\
\midrule
CircuitNet & \textbf{11.6$\pm$0.3} & 15.8$\pm$1.1 & 0.44$\pm$0.11 & 0.60$\pm$0.20 \\
AirfRANS   & \textbf{7.1$\pm$1.5}  & 12.7$\pm$1.2 & 0.30$\pm$0.24 & 0.30$\pm$0.24 \\
\bottomrule
\end{tabular}
\end{table}

\blue{Table~\ref{tab:crt} reports the resulting comparison between CRT and our TACR. At the same nominal coverage target and under the matched training protocol, \method{} has a lower mean \Sim{} on both datasets, while the observed \Miss{} values remain below one percent. This comparison complements the specialized direct-objective ablation above by evaluating a general end-to-end conformal risk-control framework adapted to the routing task.}

\section{\blue{Comparison with calibrated NEEDS gating}}
\label{app:uddin}

\blue{The calibrated gate of \citet{uddin2026proactive} and \method{} target different risks: the former controls the violation rate conditional on assigning an input to the surrogate, whereas \method{} controls the unconditional joint event of an incorrect surrogate decision escaping simulation. We therefore compare routing efficiency at matched empirical miss rates. Our strengthened Uddin-inspired baseline uses pooled features from full-$E_1+E_2$ Standard CP surrogates, replaces logistic regression with a two-hidden-layer nonlinear MLP, is fitted on a held-out calibration split, and is trained to convergence. We use ``Uddin-inspired'' because the original method operates on input features, whereas our adaptation uses surrogate features. For each seed, we sweep the gate threshold and report the lowest empirical \Sim{} whose \Miss{} does not exceed that of \method{}.}

\begin{table}[h!]
\centering
\caption{\textbf{A strengthened Uddin-inspired gate requires substantially more routing at matched empirical miss.} Values are mean$\pm$std over five seeds. The gate uses a nonlinear head trained on held-out data with features from full-$E_1+E_2$ surrogates.}
\label{tab:uddin-strengthened}
\small
\setlength{\tabcolsep}{4pt}
\begin{tabular}{lccc}
\toprule
\textbf{Dataset} & \textbf{\method{} \Sim{} / \Miss{}} & \textbf{Gate \Sim{}} & \textbf{Gate \Miss{}} \\
\midrule
AhmedML & \textbf{24.8$\pm$4.3} / 1.19$\pm$0.97 & 71.9$\pm$13.1 & 0.99$\pm$0.00 \\
ShapeNet-Car & \textbf{35.3$\pm$2.9} / 0.56$\pm$0.50 & 82.0$\pm$5.6 & 0.56$\pm$0.00 \\
\bottomrule
\end{tabular}
\end{table}

\blue{At no higher observed miss rate, the strengthened gate routes approximately $2.9\times$ as many AhmedML inputs and $2.3\times$ as many ShapeNet-Car inputs as \method{}.}

\section{Full per-dataset metrics}
\label{app:full}

Table~\ref{tab:metric-glossary} summarizes the core notation and gives plain-language interpretations of the reported metrics. \Sim{} and \Miss{} form the primary cost--risk pair, whereas $g$-MAE and \textsc{Needs}\% diagnose predictor accuracy and the surrogate-specific zero-miss routing reference, respectively.

\begin{table}[!h]
\centering
\caption{Core notation and evaluation metrics. Each quantity is paired with its operational role in conformal routing; percentages are computed over test inputs.}
\label{tab:metric-glossary}
\small
\setlength{\tabcolsep}{5pt}
\begin{tabular}{@{}p{0.13\linewidth}p{0.30\linewidth}p{0.49\linewidth}@{}}
\toprule
\textbf{Notation} & \textbf{Definition} & \textbf{Description} \\
\midrule
$\gh(x)$ & $g(f_\theta(x))$ & Surrogate estimate of the scalar decision quantity used for routing. \\
$\tau$ & Fixed decision threshold & Application-specified limit against which the decision quantity is compared, such as a maximum allowable stress or a minimum required lift coefficient. \\
$\sigma(x)$ & Positive input-dependent scale & Learned relative-width factor that allows interval width to vary across inputs. \\
$\qh$ & Calibration multiplier in Eq.~\ref{eq:conformal-quantile} & Value computed from held-out calibration scores to obtain the conformal coverage guarantee. \\
$C(x)$ & \shortstack[l]{$[\gh(x)-\qh\sigma(x),$\\$\gh(x)+\qh\sigma(x)]$} & Calibrated prediction interval for the scalar decision quantity. \\
\midrule
\Sim{} & $100\Pr(r(X)=1)$ & Inputs that still require a high-fidelity simulator call; the primary cost metric. \\
\Miss{} & \shortstack[l]{$100\Pr(\hat d(X)\neq d(Y),$\\$r(X)=0)$} & Wrong surrogate threshold decisions that escape simulation; the accompanying risk metric. \\
$g$-MAE & $\mathbb E[|\gh(X)-g(Y)|]$ & Average error in the scalar decision quantity; a diagnostic of predictor accuracy. \\
\textsc{Needs}\% & $100\Pr(\hat d(X)\neq d(Y))$ & Inputs the surrogate alone would classify incorrectly; for a fixed surrogate, the zero-miss oracle routing rate. \\
\bottomrule
\end{tabular}
\end{table}

Table~\ref{tab:full} reports the complete metric set for every method on every dataset (mean over five seeds for non-ensemble methods; ensemble single run). Precision, recall, and F1 score (Prec, Rec, and F1) evaluate routing as a detector of the \textsc{Needs}\% event, and $\rho$ is the Spearman correlation between $\sigma$ and absolute prediction error. The training objective does not directly optimize this correlation.

\begin{table}[h]
\caption{Uniform and \method{} have closely matched $g$-MAE and Needs\% across datasets, making their routing differences informative about threshold-local interval allocation rather than predictor quality. Prec, Rec, and F1 denote routing precision, recall, and F1 score. Needs, \Sim{}, and \Miss{} are percentages, whereas Prec, Rec, F1, and $\rho$ are fractions. Bold marks the best value within each dataset and metric: lower for $g$-MAE, Needs, \Sim{}, and \Miss{}, and higher for Prec, Rec, F1, and $\rho$. Displayed ties are all bolded.}
\label{tab:full}
\begin{center}
\scriptsize
\begin{tabular}{llrrrrrrrr}
\toprule
\textbf{Dataset} & \textbf{Method} & $g$-MAE & Needs\% & \Sim{} & Prec & Rec & F1 & \Miss{} & $\rho$ \\
\midrule
\multirow{6}{*}{CircuitNet}
 & Standard CP   & 0.0082 & 8.1 & 46.4 & 0.17 & 0.95 & 0.28 & 0.37 & 0.00 \\
 & CQR           & 0.0127 & 14.5 & 65.9 & \textbf{0.22} & \textbf{0.99} & \textbf{0.36} & \textbf{0.09} & -0.06 \\
 & Uniform       & \textbf{0.0023} & \textbf{2.4} & 15.3 & 0.14 & 0.90 & 0.24 & 0.23 & 0.18 \\
 & \twuq{}    & 0.0053 & 4.5 & 22.5 & 0.19 & 0.97 & 0.32 & 0.12 & \textbf{0.66} \\
 & Deep Ensemble & 0.0083 & 7.9 & 50.5 & 0.15 & \textbf{0.99} & 0.27 & 0.12 & 0.29 \\
 & \method{} & 0.0024 & \textbf{2.4} & \textbf{11.6} & 0.17 & 0.82 & 0.28 & 0.44 & 0.22 \\
\midrule
\multirow{6}{*}{Darcy}
 & Standard CP   & 0.0135 & 1.8 & 14.8 & 0.12 & 0.98 & 0.21 & 0.03 & 0.00 \\
 & CQR           & 0.0131 & 1.5 & 12.5 & 0.14 & 0.93 & 0.23 & 0.11 & 0.60 \\
 & Uniform       & 0.0129 & \textbf{1.4} & 14.1 & 0.09 & 0.99 & 0.17 & 0.02 & -0.06 \\
 & \twuq{}    & 0.0152 & 2.1 & 13.6 & 0.14 & 0.94 & 0.25 & 0.13 & 0.43 \\
 & Deep Ensemble & \textbf{0.0127} & \textbf{1.4} & \textbf{8.3} & \textbf{0.17} & \textbf{1.00} & \textbf{0.29} & \textbf{0.00} & \textbf{0.70} \\
 & \method{} & 0.0129 & 1.5 & 12.7 & 0.11 & 0.96 & 0.20 & 0.06 & -0.31 \\
\midrule
\multirow{6}{*}{AirfRANS}
 & Standard CP   & 0.0777 & 3.2 & 19.8 & 0.15 & 0.95 & 0.26 & 0.20 & 0.00 \\
 & CQR           & 0.0678 & 4.3 & 19.1 & 0.21 & 0.93 & 0.34 & 0.30 & \textbf{0.52} \\
 & Uniform       & \textbf{0.0368} & 2.9 & 9.7 & \textbf{0.29} & 0.95 & \textbf{0.44} & 0.10 & 0.25 \\
 & \twuq{}    & 0.0485 & \textbf{2.2} & 10.5 & 0.17 & 0.68 & 0.27 & 0.40 & 0.25 \\
 & Deep Ensemble & 0.0694 & 3.5 & 28.0 & 0.12 & \textbf{1.00} & 0.22 & \textbf{0.00} & 0.28 \\
 & \method{} & 0.0369 & \textbf{2.2} & \textbf{7.4} & 0.25 & 0.81 & 0.38 & 0.40 & 0.14 \\
\midrule
\multirow{6}{*}{AhmedML}
 & Standard CP   & 0.0172 & 11.7 & 31.7 & \textbf{0.31} & 0.83 & \textbf{0.45} & 1.98 & 0.00 \\
 & CQR           & 0.0153 & \textbf{9.3} & 27.9 & 0.27 & 0.83 & 0.40 & \textbf{1.78} & 0.06 \\
 & Uniform & 0.0149 & 10.1 & \textbf{25.5} & \textbf{0.31} & 0.74 & 0.43 & 2.57 & 0.00 \\
 & \twuq{}    & 0.0208 & 11.9 & 36.6 & 0.28 & \textbf{0.85} & 0.42 & \textbf{1.78} & \textbf{0.33} \\
 & Deep Ensemble & 0.0190 & 10.9 & 29.7 & 0.27 & 0.73 & 0.39 & 2.97 & 0.18 \\
 & \method{}     & \textbf{0.0148} & 9.7 & 25.7 & \textbf{0.31} & 0.75 & 0.42 & 2.38 & 0.00 \\
\midrule
\multirow{6}{*}{ShapeNet-Car}
 & Standard CP   & 0.0132 & 9.4 & 46.4 & 0.19 & 0.92 & 0.31 & 0.78 & 0.00 \\
 & CQR           & 0.0110 & 8.4 & 38.3 & 0.20 & 0.92 & 0.33 & 0.67 & 0.00 \\
 & Uniform       & 0.0101 & 8.4 & 36.4 & \textbf{0.22} & 0.93 & \textbf{0.35} & 0.56 & 0.00 \\
 & \twuq{}    & 0.0140 & 10.6 & 48.2 & 0.21 & \textbf{0.96} & \textbf{0.35} & \textbf{0.45} & \textbf{0.21} \\
 & Deep Ensemble & 0.0129 & 10.6 & 48.6 & 0.21 & 0.95 & 0.34 & 0.56 & 0.20 \\
 & \method{} & \textbf{0.0095} & \textbf{8.2} & \textbf{34.7} & \textbf{0.22} & 0.92 & \textbf{0.35} & 0.56 & 0.00 \\
\midrule
\multirow{6}{*}{Beams}
 & Standard CP   & 0.1314 & 2.8 & 9.4 & 0.28 & \textbf{0.93} & 0.43 & \textbf{0.20} & 0.00 \\
 & CQR           & 0.1302 & \textbf{2.6} & 7.8 & 0.31 & 0.92 & 0.46 & \textbf{0.20} & \textbf{0.55} \\
 & Uniform       & 0.1320 & 2.7 & 7.5 & 0.32 & 0.87 & 0.46 & 0.35 & \textbf{0.55} \\
 & \twuq{}    & 0.1329 & 2.9 & 7.2 & 0.35 & 0.86 & 0.49 & 0.40 & \textbf{0.55} \\
 & Deep Ensemble & \textbf{0.1286} & 2.8 & 14.2 & 0.18 & 0.91 & 0.29 & 0.25 & 0.37 \\
 & \method{} & 0.1336 & 2.9 & \textbf{5.7} & \textbf{0.40} & 0.79 & \textbf{0.53} & 0.60 & 0.44 \\
\bottomrule
\end{tabular}
\end{center}
\end{table}

\begin{figure}[t]
\centering
\includegraphics[width=0.55\linewidth]{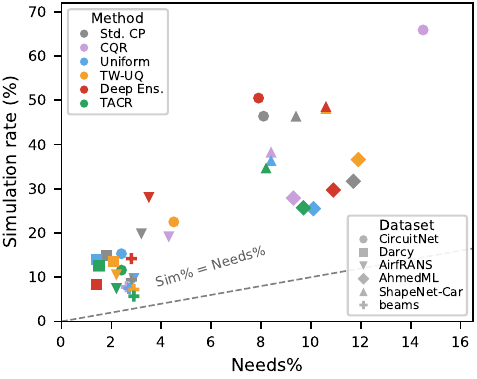}
\caption{\textbf{Routing relative to the zero-miss reference across all methods.} The dashed line, $\textsc{Sim}\%=\textsc{Needs}\%$, marks the minimum routing rate under zero-miss oracle routing for each fixed surrogate. Color denotes method, and marker shape denotes dataset; ``Std. CP'' and ``Deep Ens.'' denote Standard CP and Deep Ensemble.}
\label{fig:needs-full}
\end{figure}

\section{Conditional behavior near $\tau$}
\label{sec:conditional}
\begin{figure}[!ht]
\centering
\includegraphics[width=\linewidth]{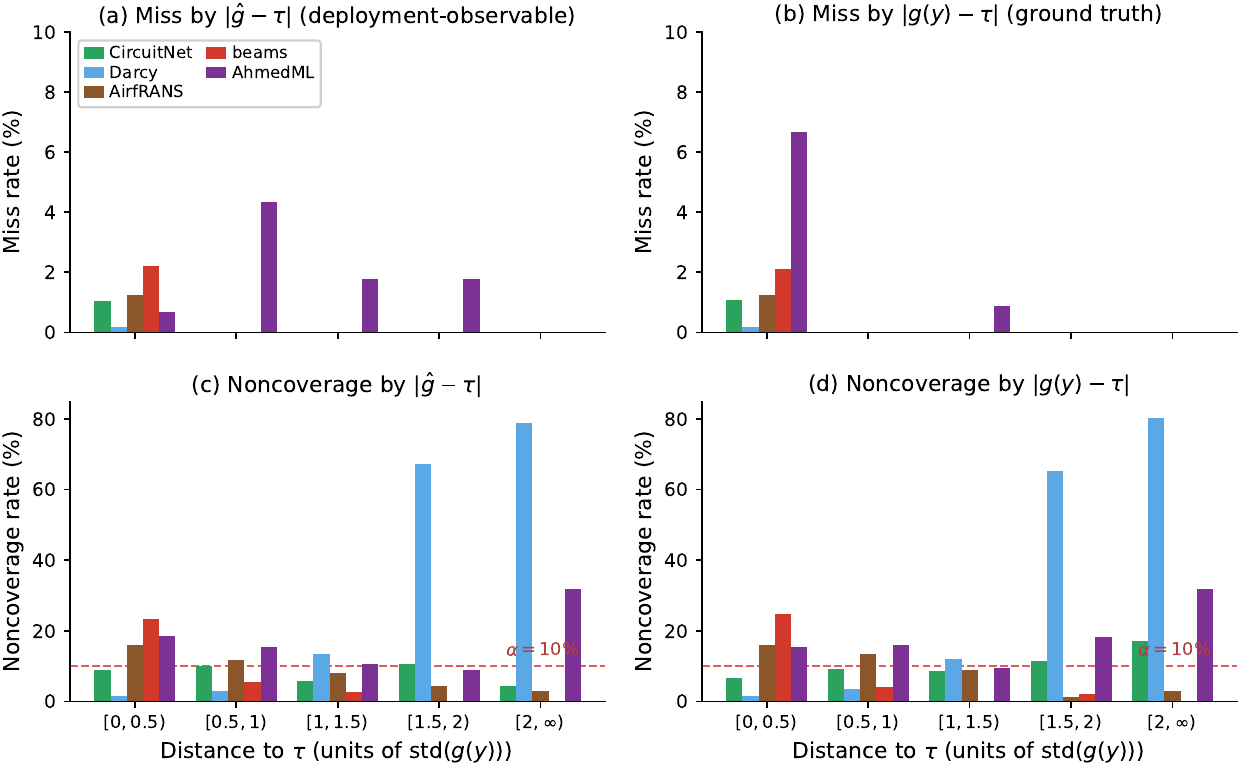}
\caption{\textbf{Distance-stratified misses and noncoverage for \method{}.} Bars pool five seeds for each displayed dataset. The top row reports miss and the bottom row noncoverage ($100\%-\mathrm{coverage}$); columns bin by the predicted margin (left) or ground-truth distance (right), normalized by the standard deviation of $g(y)$. The dashed line marks the marginal $\alpha=0.1$ noncoverage target, not a bandwise constraint.}
\label{fig:conditional}
\end{figure}

Although split conformal calibration controls marginal noncoverage, routing depends on where that noncoverage occurs. Near $\tau$, tighter intervals are less likely to cross the threshold and can reduce \Sim{}, particularly when the predicted decision quantity is accurate; noncoverage there, however, is also the most likely to produce a wrong unrouted decision. Far from $\tau$, noncoverage is often decision-benign because $g(y)$ can fall outside the interval while remaining on the same side of the threshold as $\gh$. Thus, allocating noncoverage to near-threshold or far-threshold inputs can respectively improve routing efficiency or use the error budget where it is less likely to increase \Miss{}. Figure~\ref{fig:conditional} examines this behavior by stratifying test samples according to either the predicted margin $|\gh-\tau|$ or the ground-truth distance $|g(y)-\tau|$.

The bars show that noncoverage is not distributed uniformly across distance bands: it is often elevated near the threshold or in the farthest bands relative to intermediate distances. Misses, by contrast, are concentrated near $\tau$ and vanish in most farther bands. Darcy provides the clearest example of the second effect: far-band noncoverage reaches approximately 65--80\%, yet no misses are observed there because the uncovered targets remain on the correct side of the threshold. This separation between noncoverage and decision error is consistent with the intended allocation of interval width: the resulting allocation can trade local coverage behavior for reduced routing where the decision is difficult, while noncoverage far from $\tau$ is comparatively inexpensive in decision risk. These stratified results are diagnostic rather than a conditional guarantee; applications requiring coverage within prespecified regions would need group-conditional or another locally valid calibration procedure.

\section{Implementation details}
\label{app:impl}

\textbf{Data splits.}
Table~\ref{tab:data-splits} reports the exact splits used in all finalized experiments. Each dataset is partitioned into training, calibration, and test sets, with no separate validation set. The split is generated once with seed 42 and held fixed across run seeds, so repeated runs differ only in model initialization and training stochasticity rather than data membership. Models are trained for fixed epoch counts and evaluated at the final epoch, without early stopping or checkpoint selection. The calibration partition is reserved exclusively for conformal calibration.

\begin{table}[h!]
\centering
\caption{Exact data splits used in the experiments. A dash indicates that no separate validation partition is used.}
\label{tab:data-splits}
\small
\setlength{\tabcolsep}{6pt}
\setlength{\aboverulesep}{0.2ex}
\setlength{\belowrulesep}{0.2ex}
\begin{tabular*}{\linewidth}{@{\extracolsep{\fill}}lrrrrr@{}}
\toprule
\textbf{Dataset} & \textbf{Total} & \textbf{Train} & \textbf{Validation} & \textbf{Calibration} & \textbf{Test} \\
\midrule
CircuitNet & 3,456 & 2,073 & -- & 519 & 864 \\
Darcy & 10,000 & 6,000 & -- & 2,000 & 2,000 \\
AirfRANS & 1,000 & 600 & -- & 200 & 200 \\
AhmedML & 498 & 298 & -- & 99 & 101 \\
ShapeNet-Car & 888 & 532 & -- & 177 & 179 \\
Beams & 2,000 & 1,200 & -- & 400 & 400 \\
\bottomrule
\end{tabular*}
\end{table}

\textbf{Dataset preprocessing.}
No cases are filtered from CircuitNet or Darcy. Darcy's inputs and targets are standardized. AirfRANS subsamples 5,000 surface and volume points per geometry and constructs radius graphs with radius 0.05 and at most 64 neighbors. AhmedML uses 48,000 points per geometry during training and 64,000 during calibration and testing. ShapeNet-Car uses point coordinates and surface normals, standardized across the dataset. The beams dataset is generated with seed 42 and contains five input features.

\textbf{Backbones and scale heads.}
Table~\ref{tab:dataset-hyperparameters} summarizes the dataset-specific architectures and optimization settings. CircuitNet uses ResNet18-UNet++ with its encoder initialized from ImageNet weights; all other backbones are initialized from scratch. AirfRANS uses three GraphSAGE layers with input dimension 7 and hidden, latent, and scale-projection dimensions of 64. Beams uses a three-layer MLP with width 64. AhmedML and ShapeNet-Car use GeoTransolver with hidden dimension 160, 20 layers, 128 slices, and eight heads.

For Uniform, \twuq{}, and \method{}, we parameterize
$\sigma(x)=\sigma_\phi(h_\theta(x))$,
where $h_\theta(x)$ is the predictor's penultimate backbone representation. The scale head projects the spatial, node, or latent features to a lower-dimensional representation, pools them into one vector per sample, and applies a small MLP with a Softplus output to ensure positivity. It attaches to the encoder bottleneck for convolutional models, pools node or point features for graph- and transformer-based models, and uses the shared MLP trunk for beams. Thus, the scale head reuses features already computed for prediction and introduces no separate feature extractor.

\begin{table}[h!]
\centering
\caption{Dataset-specific configurations. Batch sizes are shown as training/calibration/test. $E_1$ and $E_2$ denote the two stages used by Uniform, \twuq{}, and \method{}; single-stage methods train for $E_1$ epochs. All optimizers use weight decay $10^{-4}$ and cosine learning-rate decay within each stage. For AhmedML, the two learning rates denote \method{} and the baselines, respectively.}
\label{tab:dataset-hyperparameters}
\small
\setlength{\tabcolsep}{3.5pt}
\setlength{\aboverulesep}{0.2ex}
\setlength{\belowrulesep}{0.2ex}
\begin{tabular*}{\linewidth}{@{\extracolsep{\fill}}llclcc@{}}
\toprule
\textbf{Dataset} & \textbf{Backbone} & \textbf{Batch} & \textbf{Optimizer} & \textbf{Base LR} & \textbf{$E_1/E_2$} \\
\midrule
CircuitNet & ResNet18-UNet++ & 16/32/32 & AdamW & $10^{-4}$ & 200/100 \\
Darcy & 5-level U-Net & 32/64/64 & AdamW & $10^{-3}$ & 500/250 \\
AirfRANS & 3-layer GraphSAGE & 16/16/16 & AdamW & $10^{-3}$ & 200/100 \\
AhmedML & GeoTransolver & 4/2/2 & Adam & $10^{-3}/5\!\times\!10^{-4}$ & 300/150 \\
ShapeNet-Car & GeoTransolver & 4/2/2 & Adam & $5\!\times\!10^{-4}$ & 300/150 \\
Beams & 3-layer MLP & 64/256/256 & AdamW & $10^{-3}$ & 200/100 \\
\bottomrule
\end{tabular*}
\end{table}

\textbf{Prediction objectives.}
For Standard CP, Uniform, and \method{}, the prediction objective is mean-squared error on the predicted field for CircuitNet, Darcy, AirfRANS, AhmedML, and ShapeNet-Car, and mean-squared error on the scalar $\log_{10}$ peak-stress target for beams. CQR replaces the standard prediction head with two conditional-quantile outputs and optimizes the pinball loss. Each Deep Ensemble member uses the corresponding dataset-level prediction objective.

\textbf{\twuq{} objective.}
This baseline augments the dataset-level prediction loss with a threshold-weighted error on the scalar decision quantity and a Gaussian negative-log-likelihood objective for a positive scale $\sigma_i$ produced by a Softplus head. Let $g_i=g(y_i)$ and $\epsilon=10^{-8}$, and define
\begin{equation}
w_i^{\mathrm{TW}}=1+2\exp\!\left(-\frac{(\operatorname{sg}[\gh_i]-\tau)^2}{2\,\mathrm{bw}^2}\right),
\end{equation}
where $\operatorname{sg}$ denotes stop-gradient. During joint training, the minibatch objective is
\begin{align}
\mathcal L_{\mathrm{TW\text{-}UQ}}&=\mathcal L_{\mathrm{pred}}+\tfrac12\mathcal L_{g,\mathrm{IW}}+\tfrac12\mathcal L_{\mathrm{NLL}}, \\
\mathcal L_{g,\mathrm{IW}}&=\frac{1}{|\mathcal B|}\sum_{i\in\mathcal B}w_i^{\mathrm{TW}}(\gh_i-g_i)^2, \\
\mathcal L_{\mathrm{NLL}}&=\frac{1}{|\mathcal B|}\sum_{i\in\mathcal B}\left[\log(\sigma_i+\epsilon)+\frac{\operatorname{sg}[g_i-\gh_i]^2}{2\sigma_i^2+\epsilon}\right].
\end{align}
The stopped residual prevents the likelihood term from directly updating the mean prediction, while the weighted scalar error emphasizes prediction accuracy near $\tau$. The predictor-only warm-up uses $\mathcal L_{\mathrm{pred}}$ alone. At calibration and deployment, \twuq{} uses $s_i=|g_i-\gh_i|/(\sigma_i+\epsilon)$ and the same split-conformal interval $[\gh_i-\qh\sigma_i,\gh_i+\qh\sigma_i]$ as the other learned-scale methods. The objective and its coefficients are shared across all six datasets.

\textbf{Decision quantities on the CFD datasets.}
For AirfRANS, AhmedML, and ShapeNet-Car, the routing quantity is computed by integrating the predicted surface pressure. AirfRANS uses the lift coefficient $C_L$, whereas AhmedML and ShapeNet-Car use the drag coefficient $C_D$. ShapeNet-Car additionally provides surface normals, which are required for the drag calculation.

\textbf{Training-time soft quantile.}
Within each training mini-batch, $\operatorname{SoftQuantile}$ is implemented as a linearly interpolated empirical quantile of the normalized scores. After sorting the $m=|\mathcal B|$ scores, we evaluate the quantile at the continuous zero-based rank $(1-\alpha)(m-1)$ by interpolating between the adjacent order statistics. This approximation is used only during optimization and is distinct from the exact rank-based calibration rule in Eq.~\ref{eq:conformal-quantile}.

\blue{We test sensitivity to this approximation by holding the optimizer batch, routing-loss evaluation set, learning rates, and training schedule fixed while varying only the fraction of batch scores used to estimate the training-time quantile.}

\begin{table}[h!]
\centering
\caption{Sensitivity to the training-time quantile approximation. The quantile pool is the fraction of batch scores used to estimate $\tilde q_{\mathcal B}$, while the routing loss is evaluated on the full batch. Values are five-seed means in percent.}
\label{tab:qpool}
\small
\setlength{\tabcolsep}{6pt}
\begin{tabular}{lrrrr}
\toprule
\textbf{Dataset} & \textbf{Pool} & \textbf{\Sim{} $\downarrow$} & \textbf{\Miss{} $\downarrow$} & \textbf{Coverage} \\
\midrule
CircuitNet & 25\%  & 11.7 & 0.44 & 91.3 \\
CircuitNet & 50\%  & 10.9 & 0.32 & 90.8 \\
CircuitNet & 100\% & 11.6 & 0.44 & 90.9 \\
AirfRANS   & 25\%  & 7.1  & 0.50 & 90.4 \\
AirfRANS   & 50\%  & 7.5  & 0.20 & 90.9 \\
AirfRANS   & 100\% & 7.1  & 0.30 & 90.8 \\
\bottomrule
\end{tabular}
\end{table}

\blue{Across the tested pool fractions, \Sim{} ranges from $10.9$--$11.7\%$ on CircuitNet and $7.1$--$7.5\%$ on AirfRANS, with no monotonic degradation as the pool is reduced. Coverage remains close to the $90\%$ target and \Miss{} remains at or below $0.50\%$. Thus, routing performance is robust to the training-time approximation over the tested range. Final conformal validity is unaffected by this approximation because deployment uses exact calibration on independent held-out data.}

\textbf{Conformal calibration.}
We use $\alpha=0.1$, corresponding to 90\% target marginal coverage. After training, the conformal multiplier is computed on the held-out calibration set using the exact order statistic in Eq.~\ref{eq:conformal-quantile}. The calibration set is disjoint from both training and testing and is not used for model fitting or for the training-time soft quantile.

\textbf{Hyperparameter selection.}
All routing-objective hyperparameters are fixed globally rather than tuned separately for individual datasets. We use $\lambda=0.5$, $\beta=16$, and $\mu=10^{-2}$ for every dataset and seed, and set
$\mathrm{bw}=0.3\cdot\operatorname{std}_{\mathrm{train}}(g(y))$,
where the standard deviation is computed over the training labels. These values are heuristic defaults and are not selected through dataset-specific search or by calibration- or test-set performance. The $\beta$ sweep in Table~\ref{tab:beta} is reported only as a sensitivity analysis and is not used to select the default setting. Uniform follows the same architecture, scale-head parameterization, routing objective, and training schedule as \method{}, with the sole change $\beta=0$ to remove threshold-local weighting.

Table~\ref{tab:shared-hyperparameters} summarizes the settings shared across the main experiments.

\begin{center}
\captionof{table}{Shared settings for the main experiments. The loss coefficients describe \method{}, and Uniform uses the same routing objective with $\beta=0$. The two-stage schedule applies to Uniform, \twuq{}, and \method{}. Let $E$ and $\eta$ denote the dataset-specific base epoch count and learning rate.}
\label{tab:shared-hyperparameters}
\small
\setlength{\tabcolsep}{4pt}
\setlength{\aboverulesep}{0.2ex}
\setlength{\belowrulesep}{0.2ex}
\begin{tabular*}{\linewidth}{@{\extracolsep{\fill}}ll@{}}
\toprule
\textbf{Setting} & \textbf{Value} \\
\midrule
Conformal miscoverage level $\alpha$ & $0.1$ \\
Decision threshold $\tau$ & Training-set 70th percentile of $g(y)$ \\
Routing-loss weight $\lambda$ & $0.5$ \\
Threshold-local weighting strength $\beta$ & $16$ \\
Scale-regularization weight $\mu$ & $0.01$ \\
Bandwidth $\mathrm{bw}$ & $0.3\cdot\operatorname{std}_{\mathrm{train}}(g(y))$ (population std) \\
Stage 1 duration / predictor learning rate & $E$ / $\eta$ \\
Stage 2 duration & $E/2$ \\
Stage 2 predictor / scale-head learning rates & $0.1\eta$ / $\eta$ \\
Evaluation epoch & Final epoch \\
Non-ensemble random seeds & $42, 123, 456, 789, 1011$ \\
Deep Ensemble members & $5$ (one ensemble run) \\
\bottomrule
\end{tabular*}
\end{center}

\textbf{Training schedules.}
Uniform, \twuq{}, and \method{} use the same two-stage schedule. Stage 1 optimizes the predictor alone for the dataset-specific base duration $E$. Stage 2 jointly fine-tunes the predictor and the method-specific scale head for $E/2$ epochs, using one tenth of the base learning rate for predictor parameters and the base learning rate for scale-head parameters.

Standard CP and CQR use a single stage of $E$ epochs. CQR replaces the prediction head with two conditional-quantile outputs and optimizes the pinball loss. Each Deep Ensemble member is trained independently for $E$ epochs, using member seeds 100--104. Standard CP uses absolute residuals as nonconformity scores, CQR conformalizes its learned quantile interval, and Uniform, \twuq{}, and \method{} use the positive Softplus scale head described above.

\end{document}